\documentclass[times, review, 10pt]{elsarticle}
\usepackage{amsthm, amssymb, amsmath, amsfonts}
\usepackage{algorithmic}
\usepackage{algorithm}
\usepackage{array}
\usepackage[caption=false,font=normalsize,labelfont=sf,textfont=sf]{subfig}
\usepackage{textcomp}
\usepackage{stfloats}
\usepackage{url}
\usepackage{verbatim}
\usepackage{graphicx}
\usepackage[hidelinks,bookmarks=false]{hyperref}
\usepackage{tabularx}
\usepackage[nameinlink,noabbrev]{cleveref}
\usepackage{xr}
\usepackage{multirow}
\hypersetup{hypertex=true,
	colorlinks=true,
	linkcolor=blue,
	anchorcolor=blue,
	citecolor=blue}
\numberwithin{equation}{section}
\theoremstyle{plain}
\newtheorem{theorem}{Theorem}
\newtheorem{lemma}[theorem]{Lemma}

\newtheorem{corollary}[theorem]{Corollary}
\theoremstyle{definition}

\theoremstyle{remark}

\newcommand{\E}{\mathbb{E}}
\newcommand{\R}{\mathbb{R}}
\newcommand{\prior}{\pi}
\newcommand{\talpha}{\alpha}
\newcommand{\alphabar}{\bar{\alpha}}
\newcommand{\pU}{p_{U}}
\newcommand{\pT}{\tilde p_{T}}
\newcommand{\pp}{p_{+}}
\newcommand{\pn}{p_{-}}
\newcommand{\risk}{\mathcal{R}}

\newcommand{\pHolm}{p_{\mathrm{Holm}}}
\newcommand{\AP}{\mathrm{AP}}
\newcommand{\AUROC}{\mathrm{AUROC}}

\newcommand{\vx}{\boldsymbol{x}}

\newcommand{\ECE}{\mathrm{ECE}}
\newcommand{\Dalpha}{\Delta_{\alpha}}
\newcommand{\Dop}{\Delta_{\mathrm{op}}}
\newcommand{\dprior}{\delta}

\newcommand{\piTilde}{\tilde{\pi}}
\newcommand{\Brier}{\mathrm{Brier}}

\newcommand{\Cliff}{\delta_{\mathrm{Cliff}}}
\theoremstyle{definition}
\theoremstyle{remark}
\newcommand{\pTtil}{\tilde p_{T}}
\journal{Pattern Recognition}

\begin{document}

\begin{frontmatter}

\title{Learning from \textit{N}-Tuple Data with \textit{M} Positive Instances: Unbiased Risk Estimation and Theoretical Guarantees}

\author{Miao Zhang}
\ead{zhangmiao@stumail.ysu.edu.cn}

\author{Junping Li\corref{cor1}}
\ead{jpl@ysu.edu.cn}

\author{Changchun Hua}
\ead{cch@ysu.edu.cn}

\author{Yanyan Yang}
\ead{yyn@ysu.edu.cn}

\cortext[cor1]{Corresponding author.}

\affiliation{
	organization={Engineering Research Center of the Ministry of Education
		for Intelligent Control System and Intelligent Equipment, Yanshan University},
	city={Qinhuangdao},
	country={China}
}
\begin{abstract}
Weakly supervised learning often operates with coarse aggregate signals rather than instance labels. We study a setting where each training example is an $n$-tuple containing exactly $m$ positives, while only the count $m$ per tuple is observed. This NTMP (N-tuple with M positives) supervision arises in, e.g., image classification with region proposals and multi-instance measurements. We show that tuple counts admit a trainable unbiased risk estimator (URE) by linking the tuple-generation process to latent instance marginals. Starting from fixed $(n,m)$, we derive a closed-form URE and extend it to variable tuple sizes, variable counts, and their combination. Identification holds whenever the effective mixing rate is separated from the class prior $\pi$ (i.e., $\bar{\alpha}\neq \pi$). We establish generalization bounds via Rademacher complexity and prove statistical consistency with standard rates under mild regularity assumptions. To improve finite-sample stability, we introduce simple ReLU corrections to the URE that preserve asymptotic correctness. Across benchmarks converted to NTMP tasks, the approach consistently outperforms representative weak-supervision baselines and yields favorable precision--recall and $F_{1}$ trade-offs. It remains robust under class-prior imbalance and across diverse tuple configurations, demonstrating that count-only supervision can be exploited effectively through a theoretically grounded and practically stable objective.
\end{abstract}


\end{frontmatter}
\section{Introduction}
The rapid progress of modern AI has intensified the demand for large, labeled corpora. Fully supervised pipelines \cite{zhou2018brief,  9003490, 8674823} hinge on exhaustive instance-level annotation, which is costly or infeasible in sensitive domains such as healthcare and scientific experiments \cite{LU2023109861, YE2025112684}. This has motivated a rich literature on weakly supervised learning (WSL) from incomplete, inexact, or noisy supervision \cite{elkan2008learning, 9294086, 9526351}, including positive--unlabeled (PU) learning \cite{10.5555/3294771.3294931, bekker2020learning, du2014analysis, 8520779}, partial-label learning \cite{lv2020progressive, 10.1609/aaai.v33i01.33013542}, complementary-label learning \cite{CAO2022108447, ishida2019complementary, TANG2025111651, 9489374}, positive-confidence variants \cite{shinoda2020binary, 10.5555/3327345.3327492}, UU learning \cite{lu2018minimal, lu2020mitigating}, and tuple-/relation-based supervision using similarities, confidences, and comparisons \cite{ JMLR:v23:21-0946, bao2018classification, li2025binary, HUANG2026112230}.

A recurring practical theme is that aggregate labels are easier to collect than instance labels. Labels from Label Proportions (LLP) replace per-instance annotation with bag-level class proportions and can be learned via distributional constraints \cite{SHI20189, ding2017learning}. Multi-instance learning (MIL) similarly exploits bag-level signals \cite{ilse2018attention, carbonneau2018multiple}. However, a known limitation of pure LLP is non-identifiability when all bags share the same class proportion: if every bag has the identical positive rate, the constraints collapse to matching a single global mean, leaving many degenerate solutions and little instance-level supervision. By contrast, many applications provide an exact positive count per tuple (e.g., an image contains exactly $m$ positive regions among $n$ proposals). This signal retains the collection/privacy advantages of LLP while differing structurally: the tuple size $n$ and the exact count $m$ are known, but positions are unobserved.	
In light of these observations, we focus on the exact–count variant of aggregate supervision. Rather than treating it as yet another supervision type parallel to LLP/MIL, we ask: Can tuple-level counts be turned into a principled, trainable objective with finite-sample control, even when across-tuple variation is minimal? This perspective motivates the NTMP formulation below.

We study learning from $n$-tuples that contain exactly $m$ positives (NTMP), and denote the within-tuple positive proportion by $\alpha:=m/n$. Rather than focusing on supervision type alone—which conceptually overlaps with LLP/MIL—we develop a learning principle that turns tuple-count information into a tractable unbiased risk estimator (URE) with finite-sample control. Crucially, even if all tuples share the same proportion $\alpha$ (i.e., no across-tuple variation), our formulation remains identifiable and trainable provided $\alpha\neq\pi$, where $\pi$ is the population class prior. In practice this mild separation can be ensured by design (e.g., selecting $n$ or constructing tuples so that $\alpha$ does not coincide with $\pi$). We quantify sensitivity to this gap and show that stability scales with $\lvert \pi-\alpha\rvert^{-1}$; simple clamps keep the empirical objective well-behaved as $\alpha\to\pi$ (see Sec.~\ref{section5}). To operationalize learning, we pair NTMP tuples with an unlabeled reference set whose class prior is treated as known (an unlabeled-$\pi$ pool). The prior can be obtained from domain statistics or standard prior-estimation/calibration on a small validation batch; we also assess misspecification in Sec.~\ref{section5}. This pairing yields a closed-form linear system that eliminates the unknown class-conditionals and results in a URE amenable to standard ERM.

\paragraph*{Our contributions}
\begin{itemize}
	\item Unbiased risk from tuple counts + an unlabeled-$\pi$ pool. We formalize the NTMP data generation, show that flattening tuples yields a mixture with positive rate $\alpha=m/n$, and combine this with an unlabeled set of known prior to derive a closed-form linear system that removes class-conditionals from the population risk, resulting in a practical URE (no instance labels required).
	\item Variance-optimal intra-tuple weighting. Within the URE, we analyze per-tuple weightings and show that uniform instance weighting inside each tuple minimizes estimator variance under standard i.i.d.\ assumptions, providing a principled default for stochastic optimization.
	\item Generalization guarantees and consistency. Using Rademacher complexity, we establish generalization bounds and prove consistency (up to standard constants) under NTMP supervision.
	\item Practical stability corrections. As in other weak-supervision settings, the unbiased objective can be high-variance. We incorporate simple non-negative corrections (ReLU/ABS clamps) that mitigate overfitting while preserving asymptotic correctness.
\end{itemize}

Sec.~\ref{section2} reviews preliminaries and related work. Sec.~\ref{section3} introduces the NTMP model, derives the URE, presents generalization results, and gives the training algorithm with stability corrections. Sec.~\ref{section5} reports experiments and robustness analyses. Sec.~\ref{section6} concludes and discusses extensions (e.g., variable $(n,m)$ and tuple-aware architectures).

\section{Preliminaries}\label{section2}
This section fixes notation, recalls the supervised learning setup, and positions our problem within related weakly supervised paradigms.

\subsection{Supervised Learning}
We consider binary classification with feature space $\mathcal{X}\subset\R^d$ and label space $\mathcal{Y}=\{+1,-1\}$. Let $(\vx,y)\sim p(\vx,y)$ denote a generic example from an unknown distribution. Write the positive-class prior as
\[
\prior \;:=\; \mathbb{P}(y=+1)\in(0,1), 
\qquad 
1-\prior \;=\; \mathbb{P}(y=-1),
\]
and let the class-conditionals be $\pp(\vx):=p(\vx\mid y=+1)$ and $\pn(\vx):=p(\vx\mid y=-1)$. The marginal over features decomposes as
\[
p(\vx) \;=\; \prior\,\pp(\vx) + (1-\prior)\,\pn(\vx).
\]
A classifier is a scoring function $g:\mathcal{X}\to\R$ (the sign of $g(\vx)$ predicts the label). Given a loss $\ell:\R\times\{+1,-1\}\to\R_{+}$ (e.g., $0$--$1$, logistic, hinge), the population risk is
\begin{equation}\label{eq:supervised-risk}
	\begin{aligned}
		\risk(g)
		\;&=\; \E_{(\vx,y)\sim p}\big[\ell\big(g(\vx),y\big)\big] \\
		\;&=\; \prior\,\E_{\pp}\!\big[\ell\big(g(\vx),+1\big)\big]
		\;+\; (1-\prior)\,\E_{\pn}\!\big[\ell\big(g(\vx),-1\big)\big].
	\end{aligned}
\end{equation}
In fully supervised learning, one minimizes an empirical approximation of \eqref{eq:supervised-risk} using labeled samples. In our weakly supervised setting, pointwise labels are unavailable; we will express $\risk(g)$ using tuple-level count signals together with an unlabeled reference pool (Sec.~\ref{section3}).
\subsection{Related Work}
\paragraph*{UU learning}
UU learning trains a binary classifier from two unlabeled datasets whose class priors differ and are known (or estimable) \cite{lu2018minimal, lu2020mitigating}. Let $\mathcal{U}_1,\mathcal{U}_2$ be unlabeled sets drawn from mixtures with priors $\pi_1\neq\pi_2$; then $(p_{+},p_{-})$ can be recovered via a $2{\times}2$ linear system, yielding an unbiased risk estimator akin to PU learning. Our NTMP setting is distinct but connected: if we flatten all tuples into individual instances, we obtain an unlabeled pool whose effective positive rate equals the tuple ratio $\alpha:=m/n$. By coupling this pool with an additional unlabeled dataset whose prior $\pi$ is known, we obtain an analogous linear system eliminating $(p_{+},p_{-})$ and leading to a tractable unbiased risk (Sec.~\ref{section3}). This perspective places NTMP within the broader family of risk-reconstruction methods that exploit mixtures with differing priors \cite{lu2018minimal, lu2020mitigating}.

\paragraph*{Pairwise similarity/dissimilarity supervision}
Another related line learns from pairwise same-label (SU) or different-label (DU) constraints. Here the weak labels reside in
\(
\mathcal{Y}_{\mathrm{S}}=\{(+1,+1),(-1,-1)\},\quad
\mathcal{Y}_{\mathrm{D}}=\{(+1,-1),(-1,+1)\}
\)
and an unlabeled distribution is typically used to identify class marginals. NTMP formulation generalizes the pairwise cases to $n$-tuples with known positive counts $m$: when $(n,m)=(2,1)$ it recovers dissimilar pairs; when $(n,m)\in\{(2,0),(2,2)\}$ it recovers similar pairs. Unlike methods that require pair identities, orders, or confidence scores, NTMP uses only count-level constraints and still admits an unbiased-risk objective once coupled with an unlabeled reference.

\paragraph*{Learning from Label Proportions (LLP).}
LLP assumes bags of unlabeled instances with observed class proportions per bag, aiming at instance-level prediction. Early work established proportion-based formulations and consistency under identifiable mixture assumptions \cite{JMLR:v10:quadrianto09a}; max-margin variants (often termed proportion-SVM) optimize latent labels subject to proportion constraints \cite{pmlr-v28-yu13a}; other approaches cast LLP as constrained clustering/partitioning \cite{stolpe2011llp}; recent analyses view LLP through mutual contamination with identification and risk consistency \cite{NEURIPS2020_fcde1491}, alongside practical adaptations to high-dimensional data \cite{SHI20189} and adversarial training \cite{NEURIPS2019_4fc84805}. A known limitation of pure LLP is non-identifiability when all bags share the same proportion, as the constraints collapse to a single global mean. In contrast, NTMP supervises via exact positive counts in fixed/variable-size tuples and combines them with an auxiliary unlabeled pool of known prior to derive a closed-form unbiased risk (with stability corrections). Thus, while NTMP shares LLP’s motivation of aggregate supervision, its identification route and training objective are closer in spirit to UU-style risk reconstruction.
\begin{table}[t]
	\centering
	\caption{Notation used throughout the paper.}
	\label{tab:notation}
	\begin{tabularx}{\linewidth}{@{}l X@{}} 
		\hline
		Symbol & Meaning \\
		\hline
		$\prior$ & Class prior $\mathbb P(y=+1)$ \\
		$\talpha:=m/n$ & Tuple-level positive ratio (tuple length $n$, positives $m$) \\
		$\pp, \pn$ & Class-conditionals $p(x\mid y=\pm 1)$ \\
		$\pU$ & Unlabeled marginal $\prior\pp+(1-\prior)\pn$ \\
		$\pT$ & Per-instance marginal from tuples $\talpha\pp+(1-\talpha)\pn$ \\
		$g:\mathcal X\to\R$ & Scoring function / classifier \\
		$\phi(\cdot)$ & Surrogate loss; $\ell(y,g)=\phi(y\,g)$ \\
		$\risk(g)$ & Population risk under $\prior$ \\
		\hline
	\end{tabularx}
\end{table}

\section{Unbiased Risk Minimization Method}\label{section3}
We present a framework for learning from $n$-tuples that contain exactly $m$ positive instances (NTMP). We first formalize the tuple data–generation process and relate it to the underlying class–conditionals $(p_{+},p_{-})$. This yields a closed–form unbiased risk estimator (URE) that enables training without any instance–level labels. We then provide theory—identifiability, variance–aware weighting, and generalization—and describe a training algorithm equipped with a simple stability correction. For clarity, we develop the main results under a basic setting where both the tuple length $n$ and the positive count $m$ are fixed across tuples; the extensions to variable $n$, variable $m$, and jointly variable $(n,m)$ are presented in Appendix~A.	
\subsection{Data Generation}
Let $D_T=\{T_i\}_{i=1}^{n_T}$ be a dataset of tuples, where $T_i=(x_i^{(1)},\ldots,x_i^{(n)})$ and each tuple contains exactly $m$ positives and $n-m$ negatives (positions unknown). The tuple-level label space is
\[
	\mathcal{Y}_T
	=\Big\{\boldsymbol y\in\{+1,-1\}^n:\;\textstyle\sum_{j=1}^{n}\mathbf 1\{y^{(j)}=+1\}=m\Big\}, 
	\quad |\mathcal{Y}_T|=\binom{n}{m}.
\]

\begin{lemma}[Tuple density as an equally weighted mixture]\label{lem:tuple-density}
	Under the NTMP process, the joint density of a tuple factorizes as
	\[
	p_T(x^{(1)},\ldots,x^{(n)})
	= \frac{1}{\binom{n}{m}} \sum_{\boldsymbol y\in\mathcal{Y}_T}
	\prod_{j=1}^{n} p\!\big(x^{(j)}\mid y^{(j)}\big),
	\]
	where the sum is over all label assignments $\boldsymbol{y}=(y^{(1)},\dots,y^{(n)})$ with exactly $m$ entries equal to $+1$.
\end{lemma}

Flattening all tuples yields a pointwise multiset $\tilde D_T$ containing $n_T n$ instances. The full proof is given in Appendix C, see Lemmas A.7.
The following describes its marginal.
\begin{theorem}[Flattened marginal]\label{thm:flattened-marginal}
	Let $\alpha:=m/n$. If each $T_i$ contains exactly $m$ positives and tuples are generated independently with exchangeable positions, then the flattened index $(i,j)$ has marginal
	\[
	\pTtil(x) \;=\; \alpha\,\pp(x) + (1-\alpha)\,\pn(x),
	\]
	so that draws sampled uniformly (with replacement) from $\tilde D_T$ are i.i.d.\ from $\pTtil$.
\end{theorem}

Thus $\pTtil$ is a fixed two-component mixture whose weights are determined by the tuple composition. The full proof is given in Appendix C, see Theorem A.8. Let $p_U(x)=\prior\,\pp(x)+(1-\prior)\,\pn(x)$ denote the marginal of an auxiliary unlabeled dataset $D_U=\{u_i\}_{i=1}^{n_U}$, with known (or consistently estimable) prior $\prior\in(0,1)$. Combining $\pTtil$ and $p_U$ yields a $2{\times}2$ linear system in $(\pp,\pn)$.

\begin{lemma}[Identification from $(\pTtil,p_U)$]\label{lem:identification}
	If $\alpha\neq \prior$ (identifiability), then for every $x$,
	\begin{equation}\label{eq:ident}
		\begin{aligned}
			\pp(x)&=\frac{(1-\alpha)\,p_U(x)-(1-\prior)\,\pTtil(x)}{\prior-\alpha},\\
			\pn(x)&=\frac{-\alpha\,p_U(x)+\prior\,\pTtil(x)}{\prior-\alpha}.
		\end{aligned}
	\end{equation}
\end{lemma}
The full proof is given in Appendix C, see Lemma A.9.
\subsection{Unbiased Risk}\label{section3-b}
Recall the supervised population risk (cf.~\eqref{eq:supervised-risk}):
\[
\risk(g)=\prior\,\E_{\pp}\!\big[\ell(g(\vx),+1)\big]+(1-\prior)\,\E_{\pn}\!\big[\ell(g(\vx),-1)\big].
\]
Substituting \eqref{eq:ident} and using linearity of expectation expresses $\risk(g)$ via expectations under $\pTtil$ and $p_U$ only.

\begin{theorem}[Unbiased risk representation for NTMP]\label{thm:ure}
	Let $\alpha=m/n$ and assume $\alpha\neq\prior$. Then for any measurable $g$,
	\begin{equation}\label{eq:ure-pop}
		\begin{aligned}
			\risk(g)
			&= \frac{1}{\prior-\alpha}\Big[
			\prior(1-\alpha)\,\E_{p_U}\!\big[\ell(g(\vx),+1)\big] \;\;-\prior(1-\prior)\,\E_{\pTtil}\!\big[\ell(g(\vx),+1)\big] \\
			&\qquad\qquad\;\;-(1-\prior)\alpha\,\E_{p_U}\!\big[\ell(g(\vx),-1)\big]+(1-\prior)\prior\,\E_{\pTtil}\!\big[\ell(g(\vx),-1)\big]\Big].
		\end{aligned}
	\end{equation}
	Hence $\risk(g)$ can be estimated unbiasedly from samples of $\pTtil$ (tuples, flattened) and $p_U$ (unlabeled pool), without any instance labels.
\end{theorem}
The full proof is given in Appendix C, see Theorem A.10.
\paragraph*{Empirical estimator}
Let $\{x_i^{(j)}\}$ be all within-tuple instances and $\{u_i\}$ the unlabeled samples. Denote the per-tuple uniform average
\(
\frac{1}{n}\sum_{j=1}^{n} h\!\big(x_i^{(j)}\big)
\)
to estimate $\E_{\pTtil}[h(\vx)]$. The plug-in empirical URE is
\begin{equation}\label{eq:emp-risk}
	\begin{aligned}
		\widehat{\risk}(g)
		= \frac{1}{\prior-\alpha}\Bigg[
		\prior(1-\alpha)\,\frac{1}{n_U}\sum_{i=1}^{n_U}\ell\!\big(g(u_i),+1\big)
		\;-\prior(1-\prior)\,\frac{1}{n_T}\sum_{i=1}^{n_T}\frac{1}{n}\sum_{j=1}^{n}\ell\!\big(g(x_i^{(j)}),+1\big) \\
		\;-( 1 - \prior) \frac{\alpha}{n_U} \sum_{i=1}^{n_U}\ell \big( g(u_i), -1 \big) + ( 1 - \prior )\,\frac{\prior}{nn_T} \sum_{i=1}^{n_T}\sum_{j=1}^{n}\ell \big(g(x_i^{(j)} ),-1\big)
		\Bigg],
	\end{aligned}
\end{equation}
for which $\E\!\big[\widehat{\risk}(g)\big]=\risk(g)$ holds for all $g$. More generally, one could weight positions inside a tuple by nonnegative coefficients $(a_1,\ldots,a_n)$ with $\sum_{j}a_j=1$. The following shows the uniform choice minimizes variance.

\begin{lemma}[Uniform in-tuple averaging is variance-minimizing]\label{lem:variance}
	Among all convex combinations $\sum_{j=1}^{n} a_j h(x^{(j)})$ with $a_j\ge 0$ and $\sum_j a_j=1$, the choice $a_j=1/n$ minimizes $\mathrm{Var}\!\big(\sum_{j} a_j h(x^{(j)})\big)$ whenever the summands have common variance and nonnegative covariances (in particular, under i.i.d.\ or exchangeable symmetry within tuples). Equality holds only for the uniform weights.
\end{lemma}
The full proof is given in Appendix C, see Lemma~A.11. We only require equal marginal variances and nonnegative within-tuple covariances (i.i.d./exchangeable being special cases) for the variance-minimizing property of uniform in-tuple averaging. Independence is not required; it is used only as a convenient approximation in Appendix proofs.

\paragraph*{A stability correction (nonnegativity clamp)}\label{sec:stability-clamp}
Unbiased objectives in weak supervision can have high variance and take negative empirical values at finite sample sizes. 
We therefore use a clamped empirical objective
\[
\widehat{\mathcal{R}}_{\mathrm{corr}}(g)
\;:=\;
f\big(\widehat{\mathcal{R}}_{T}(g)\big)\;+\;f\big(\widehat{\mathcal{R}}_{U}(g)\big),
f(z)\in\{\mathrm{ReLU}(z),\,|z|\},
\]
where
\[
	\widehat{\mathcal{R}}_{T}(g):=\frac{1}{\pi-\alpha}\Big[
	-\pi(1-\pi)\,\widehat{\mathbb{E}}_{\pTtil}\big[\ell(g(\boldsymbol{x}),+1)\big] +(1-\pi)\pi\,\widehat{\mathbb{E}}_{\pTtil}\big[\ell(g(\boldsymbol{x}),-1)\big]
	\Big],
\]
\[
	\widehat{\mathcal{R}}_{U}(g):=\frac{1}{\pi-\alpha}\Big[
	\pi(1-\alpha)\,\widehat{\mathbb{E}}_{p_U}\big[\ell(g(\boldsymbol{x}),+1)\big]-(1-\pi)\alpha\,\widehat{\mathbb{E}}_{p_U}\big[\ell(g(\boldsymbol{x}),-1)\big]
	\Big],
\]
and $\widehat{\mathbb{E}}$ denotes the corresponding empirical averages (tuples flattened uniformly for $\pTtil$). 
The clamp $f$ controls variance and curbs overfitting in practice while introducing an upward bias that vanishes as sample size grows (bounded losses and concentration). 
Importantly, if the population components are nonnegative, the population minimizer of the original risk $\mathcal{R}(g)$ is preserved.

All formulas depend on the separation factor $\lvert \prior-\alpha\rvert$. When $\alpha$ approaches $\prior$, the linear system becomes ill-conditioned and variance inflates; our robustness experiments in Sec.~\ref{section5} visualize this near-degenerate regime and confirm the theoretical prediction.

\providecommand{\Dalpha}{\Delta_{\alpha}}
\providecommand{\Dop}{\Delta_{\mathrm{op}}}
\providecommand{\dprior}{\delta}
\providecommand{\alphabar}{\bar{\alpha}}
\providecommand{\piTilde}{\tilde{\pi}}
\subsection{Class-prior estimation and robustness sweep}\label{sec:prior-window}

The procedure below is used only when the unlabeled prior is unknown
or for robustness analyses in Sec.~\ref{section5};
our main experiments adopt the known-by-construction prior described in Sec.~\ref{section5}.

\paragraph{Class-prior estimation protocol}
Before training on each dataset/setup, we run a three-step protocol:
(i) NP-style lower bound to exclude pathological priors;
(ii) mixture proportion estimation (MPE) to obtain a point estimate $\hat\prior$;
(iii) a controlled  {signed prior sweep} around $\hat\prior$.

\paragraph{Three-step template with implementation options}
\begin{enumerate}
	\item NP-style lower bound (sanity gate).
	Train a lightweight score model $s(\vx)$ (e.g., a 2-layer MLP or a linear probe on frozen features) to separate a small high-precision positive proxy (e.g., top-$k$ instances from tuples with the highest predicted margins) from the unlabeled pool $U$.
	Compute the ROC curve on a held-out validation split (size: $5{,}000$ examples or $10\%$ of $U$, whichever larger).
	Define the NP lower bound
	\[
	\prior_{\mathrm{LB}}:=\max_{\tau}\{\widehat{\mathrm{TPR}}(\tau)-\widehat{\mathrm{FPR}}(\tau)\}_{+}
	\]
	with Clopper--Pearson bands.
	 {Options:} grid over $200$ thresholds; 10-fold stratified CV if $U$ is small.
	
	\item Point estimate via MPE.
	Estimate the mixture proportion by a tail-CDF ratio with kernel smoothing:
	\[
	\hat\prior=\max_{t\in\mathcal{T}} \frac{\widehat{F}_U(t)}{\widehat{F}_P(t)},
	\]
	where $\widehat{F}(\cdot)$ are monotone tail-CDFs of $s(\vx)$ (Gaussian kernels; bandwidth by Silverman's rule) post-processed by isotonic regression; $\mathcal{T}$ is a quantile grid (e.g., 200 points on $[0.6,0.99]$ of $P$'s score distribution).
	 {Uncertainty:} bootstrap $B{=}1{,}000$ resamples on the validation split for a 95\% percentile CI of $\hat\prior$.
	
	\item Signed prior sweep and reporting.
	Around $\hat\prior$, define a symmetric grid of  {signed offsets}
	\[
	\dprior\in\{-0.30,-0.28,\dots,0.30\},\qquad
	\piTilde=\hat\prior+\dprior\in(0,1).
	\]
	Retrain (or reweight) by plugging each $\piTilde$ into the URE.
	For each $\piTilde$, evaluate the primary metric $M$ (AP or Macro-F1) across $S$ seeds (default $S{=}5$) and compute mean, std, and a 95\% bootstrap CI (default $B{=}10{,}000$).
\end{enumerate}

\paragraph{Reproducible robustness window $W_{\dprior}$}
Let $M(\piTilde)$ denote the primary metric evaluated across $S$ seeds at $\piTilde=\hat\prior+\dprior$.
We define
\[
W_{\dprior}(\varepsilon,w^\star)=\Big\{\dprior:\;
\underbrace{\big|\E[M(\hat\prior)]-\E[M(\hat\prior+\dprior)]\big|\le \varepsilon}_{\text{performance drop}}
\ \wedge\
\underbrace{\mathrm{CIwidth}_{95\%}\!\big(M(\hat\prior+\dprior)\big)\le w^\star}_{\text{stability}}
\Big\},
\]
with fixed thresholds $\varepsilon{=}0.02$ (absolute metric points) and $w^\star{=}0.05$ unless noted.
We report the maximal contiguous interval $[\dprior_{\min},\dprior_{\max}]\subseteq W_{\dprior}$ and mark
\[
\delta_{\text{crit}}^{\text{CI}}
:=\min\{\,|\dprior|:\ \mathrm{CIwidth}_{95\%}(M(\hat\prior+\dprior))\ge w^\star\,\}
\]
as a red dashed line in the sweep plots.

\subsection{Conditioning and practical design via $\Dalpha=|\prior-\alpha|$}\label{sec:conditioning-design}
This section complements \Cref{sec:prior-window} by explaining why the $2{\times}2$ identification system becomes ill-conditioned as $\alpha\!\to\!\prior$ (conditioning term $\lvert\prior-\alpha\rvert^{-1}$) and by fixing concrete data-design and training rules that keep the empirical URE well behaved without altering the population minimizer.
Intuitively, when the tuple composition matches the unlabeled prior, the contributions from the tuple-induced and unlabeled marginals nearly cancel, amplifying noise; the variance of $\widehat{\mathcal{R}}$ scales with $\lvert\prior-\alpha\rvert^{-2}$ up to constants (cf. generalization bound).
In practice, we enforce a small  {operating margin}
\[
\Dop:=|\alpha-\hat\prior| \ \ge\ \varepsilon_0\quad(\text{e.g., }\varepsilon_0\in[0.05,0.10]),
\]
by mixing at least two $(n,m)$ configurations to spread $\alpha$ and—if needed—mildly reweighting/resampling the unlabeled pool so that its empirical prior stays within $\hat\prior\pm\varepsilon_0$.
During training we down-weight or skip mini-batches that violate this margin and apply the nonnegativity clamp $f\!\in\{\mathrm{ReLU}(\cdot),\lvert\cdot\rvert\}$ from Sec.~\ref{sec:stability-clamp} to reduce variance while preserving the population minimizer.

For variable tuple lengths and/or counts, let $\alpha_t := m_t/n_t$ denote the per-tuple positive mixing rate and define the effective mixing weight $\alphabar := \E[\alpha_t]$ under the sampling scheme used in training (empirically, the weighted average over tuples that enter the loss).
Together with the unlabeled prior $\prior$, the observable pair $(\E_{\text{tuple}}[\cdot],\,\E_{\text{unlab}}[\cdot])$ forms a $2{\times}2$ linear system whose unknowns are the class-conditional risks $\big(R_+(g),R_-(g)\big)$.
Whenever $\alphabar\neq \prior$, this system is invertible and yields the same unbiased risk estimator (URE) as in the fixed $(n,m)$ case by replacing $\alpha$ with $\alphabar$.
When $\alphabar=\prior$ (the ill-conditioned case), we create two strata with distinct effective rates, $\alphabar_1\neq\alphabar_2$ (e.g., by partitioning on $n_t$ or $m_t$, or by non-uniform re-sampling); solving the two $2{\times}2$ systems on each stratum restores identifiability.
We then apply the same finite-sample stabilization (ABS/ReLU clamp) as in the base method.

\begin{theorem}[Identifiability with heterogeneous tuples]\label{prop:ext-ident}
	Let $\alpha_t=m_t/n_t$ be the tuple-level positive mixing rate and
	$\alphabar:=\E[\alpha_t]$ the effective rate induced by the sampling scheme (empirically, the average over tuples used by the loss).
	Let $\prior$ be the class prior of the unlabeled pool. Consider the linear system
	linking observables to class-conditional risks:
	\[
	\begin{bmatrix}
		\alphabar & 1-\alphabar \\
		\prior & 1-\prior
	\end{bmatrix}
	\begin{bmatrix}
		R_+(g)\\ R_-(g)
	\end{bmatrix}
	=
	\begin{bmatrix}
		T(g)\\ U(g)
	\end{bmatrix},
	\qquad
	T(g):=\E_{\text{tuple}}[\ell(g)],\;
	U(g):=\E_{\text{unlab}}[\ell(g)].
	\]
	\textbf{(i)} If $\alphabar\neq\prior$, the system is invertible and the unbiased risk
	estimator is obtained by replacing the fixed-$\alpha$ coefficients with $\alphabar$.
	\textbf{(ii)} If $\alphabar=\prior$, identifiability is restored by constructing two strata
	$\mathcal{S}_1,\mathcal{S}_2$ with distinct effective rates $\alphabar_1\neq\alphabar_2$ (e.g., partition on $n_t$ or $m_t$, or apply non-uniform re-sampling), and solving the two systems on $\mathcal{S}_1$ and $\mathcal{S}_2$ separately.
	\textbf{(iii)} In practice we use the same finite-sample stabilization as the base method (ABS/ReLU clamp) after recovering $(R_+,R_-)$.
\end{theorem}
\subsection{Theoretical Analysis}
We present generalization guarantees for the NTMP learner, show consistency with optimal rates, and describe a simple stability correction. Throughout this section we assume the identifiability condition $\alpha\neq\prior$ and use the unbiased risk representation in \eqref{eq:ure-pop} with its empirical counterpart \eqref{eq:emp-risk}.

\paragraph*{Complexity measure.}
For a function class $\mathcal{G}\subset\{g:\mathcal{X}\!\to\!\R\}$ and a distribution $\mu$ on $\mathcal{X}$, we write the (distribution-dependent) Rademacher complexity with $s$ i.i.d.\ samples $X_{1:s}\!\sim\!\mu$ as
\[
\mathfrak{R}_s^{\mu}(\mathcal{G})
:=\E_{X_{1:s}}\;\E_{\sigma_{1:s}}\!\left[\;\sup_{g\in\mathcal{G}}\frac{1}{s}\sum_{i=1}^{s}\sigma_i\, g(X_i)\right],
\]
where $\sigma_i\stackrel{\text{i.i.d.}}{\sim}\mathrm{Unif}\{\pm1\}$. Assume $\phi(t):=\ell(t,+1)$ and $\psi(t):=\ell(t,-1)$ are $\rho$-Lipschitz in $t$ and that $|g(\vx)|\le C_g$ for all $g\in\mathcal{G},\vx\in\mathcal{X}$, which implies bounded losses $0\le \ell(g(\vx),y)\le C_\ell$ with $C_\ell:=\max\{\phi(C_g),\psi(C_g)\}$.

\begin{theorem}[Generalization bound and rate]\label{thm:gen}
	Let $g^\star\in\arg\min_{g\in\mathcal{G}}\risk(g)$ and $\hat g\in\arg\min_{g\in\mathcal{G}}\widehat{\risk}(g)$, where $\widehat{\risk}$ is the empirical URE in \eqref{eq:emp-risk} based on $n_T$ tuples (each of length $n$) and $n_U$ unlabeled points. Then, for any $\delta\in(0,1)$, with probability at least $1-\delta$,
	\begin{equation}\label{eq:gen-bound}
		\begin{aligned}
			\risk(\hat g)-\risk(g^\star)
			\;& \le\;
			\frac{C_1\,\rho}{\lvert \prior-\alpha\rvert}
			\!\left(
			\mathfrak{R}_{\,n_T n}^{\,\pTtil}(\mathcal{G})\;+\;\mathfrak{R}_{\,n_U}^{\,p_U}(\mathcal{G})
			\right)
			\\
			&\quad +\;
			\frac{C_2\,C_\ell}{\lvert \prior-\alpha\rvert}
			\!\left(
			\sqrt{\frac{\log\frac{2}{\delta}}{n_T n}}
			+
			\sqrt{\frac{\log\frac{2}{\delta}}{n_U}}
			\right),
		\end{aligned}
	\end{equation}
	for absolute constants $C_1,C_2>0$. In particular, if $\mathfrak{R}_s^\mu(\mathcal{G})\le C_{\mathcal{G}}/\sqrt{s}$, then
	\[
	\begin{split}
		\risk(\hat g)-\risk(g^\star)
		\;=\;
		\mathcal{O}\!\left(\frac{1}{\lvert \prior-\alpha\rvert}\right)
		\cdot
		\left(
		\frac{1}{\sqrt{n_T n}}+\frac{1}{\sqrt{n_U}}
		\right)
		\\
		\quad\text{(ignoring log factors).}
	\end{split}		\]
	Thus the estimator is consistent as $n_T,n_U\!\to\!\infty$ and achieves the optimal $N^{-1/2}$ rate with $N\!\approx\! n_T n+n_U$ effective samples.
\end{theorem}
\noindent Scaling near $\Delta\!=\!|\pi-\alpha|$.
At the population-variance level, the matrix inversion amplifies noise at rate $\propto \Delta^{-2}$; correspondingly, the estimation-error terms in the generalization bound (Eq.~\ref{eq:gen-bound}) scale as $O(\Delta^{-1})$.
This variance--vs.--bound distinction matches the empirical “ill-conditioned band’’ in Fig.~\ref{fig:robust}.

\begin{proof}[Proof sketch]
	Apply symmetrization to \eqref{eq:emp-risk} and \eqref{eq:ure-pop} separately for the $\pTtil$ and $p_U$ parts, use the contraction lemma to pass from $g$ to the Lipschitz losses $\ell(g,\pm1)$, and combine deviations via a union bound. The coefficients in \eqref{eq:ure-pop} contribute the factor $\lvert \prior-\alpha\rvert^{-1}$, and standard Hoeffding-type bounds for bounded losses yield the concentration terms. Full details are deferred to an accompanying Appendix C, see Theorem A.16.
\end{proof}
\paragraph*{Variance-optimal in-tuple averaging}
As noted after \eqref{eq:emp-risk}, replacing the in-tuple uniform average by $\sum_{j=1}^n a_j\,h(x^{(j)})$ with $\sum_{j}a_j=1$ increases variance unless $a_j\equiv 1/n$ (Lemma~\ref{lem:variance}). We therefore adopt uniform in-tuple averaging throughout to tighten the constants in \eqref{eq:gen-bound}.

\noindent Remarks.
(i) The factor $\lvert \prior-\alpha\rvert^{-1}$ in \eqref{eq:gen-bound} captures the conditioning of the $2{\times}2$ system; as $\alpha\!\to\!\prior$ the problem becomes ill-conditioned, matching the robustness study in Sec.~\ref{section5}.  
(ii) If $\phi,\psi$ are $L$-Lipschitz and $\lvert g\rvert\le G$, one may replace $C_\ell$ by $LG$ in \eqref{eq:gen-bound}.  
(iii) The bound extends to variable $(n,m)$ by conditioning on strata of $\alpha$ and averaging the corresponding deviations.  
(iv) Practical tip. To avoid the near-degenerate regime $\alpha \approx \prior$, enforce a design margin $\lvert\alpha-\hat{\prior}\rvert\!\ge\!\varepsilon$ (e.g., $\varepsilon\in[0.05,0.10]$): mix tuples from at least two $(n,m)$ configurations to spread $\alpha$, and, if needed, mildly reweight/resample the unlabeled pool so that its empirical prior stays within $\hat{\prior}\pm\varepsilon$. During training, down-weight or skip mini-batches whose current $\alpha$ violates this margin to keep the URE well-conditioned.
\subsection{Risk Bias and Excess Risk under Prior Misspecification}\label{subsec:misspec-prior}
In practice, the true class prior $\prior$ is rarely known and is replaced by an estimate $\hat\prior$ from side information or a preliminary model. Because identification of $(\pp,\pn)$ from $(p_U,\pTtil)$ relies on the $2{\times}2$ system with determinant $(\prior-\alpha)$, any prior error may be amplified when $\alpha=\tfrac{m}{n}$ is close to $\prior$. This section quantifies how such misspecification propagates to the population risk. We assume a bounded surrogate $\phi$ (e.g., logistic, ramp, hinge clipped to a finite range) and emphasize the identifiability condition $\alpha\neq\prior$.

\begin{theorem}[Risk bias under misspecified class prior]\label{thm:prior-misspec-bias}
	Let $\prior:=\mathbb{P}(y=+1)\in(0,1)$ and let $p_U=\prior\,\pp+(1-\prior)\,\pn$. In the NTMP setting with tuple length $n$ and positive count $m$, the flattened marginal is
	\[
	\pTtil \;=\; \alpha\,\pp + (1-\alpha)\,\pn,\qquad \alpha:=\frac{m}{n},
	\]
	and $\alpha\neq\prior$. Let $\ell(y,g(\vx))=\phi\!\big(y\,g(\vx)\big)$ with $0\le \phi\le B$, and define
	\[
	\risk(g)\;=\;\prior\,\E_{\pp}[\phi(g(\vx))] \;+\; (1-\prior)\,\E_{\pn}[\phi(-g(\vx))].
	\]
	Consider the plug-in representation using a misspecified prior $\hat\prior=\prior+\delta\in(0,1)$ with $\hat\prior\neq\alpha$:
	\[
	\widehat p_{+}^{(\hat\prior)}=\frac{(1-\alpha)\,p_U-(1-\hat\prior)\,\pTtil}{\hat\prior-\alpha},
	\qquad
	\widehat p_{-}^{(\hat\prior)}=\frac{-\alpha\,p_U+\hat\prior\,\pTtil}{\hat\prior-\alpha},
	\]
	and the plug-in population risk
	\[
	\widetilde{\risk}(g;\hat\prior)\;=\;\prior\,\E_{\widehat p_{+}^{(\hat\prior)}}[\phi(g(\vx))]
	\;+\; (1-\prior)\,\E_{\widehat p_{-}^{(\hat\prior)}}[\phi(-g(\vx))].
	\]
	Then for every measurable $g$,
	\[
	\big|\widetilde{\risk}(g;\hat\prior)-\risk(g)\big|
	\;\le\;
	\frac{2B\,|\delta|}{\gamma^2},
	\gamma:=\min_{\xi\in[\min\{\prior,\hat\prior\},\,\max\{\prior,\hat\prior\}]}\!|\xi-\alpha|.
	\]
	In particular, if $\prior$ and $\hat\prior$ lie on the same side of $\alpha$, then $\gamma=\min\{|\prior-\alpha|,|\hat\prior-\alpha|\}$ and
	\[
	\big|\widetilde{\risk}(g;\hat\prior)-\risk(g)\big|
	\;\le\;
	\frac{2B\,|\delta|}{\min\{|\prior-\alpha|,|\hat\prior-\alpha|\}^{2}}.
	\]
	Moreover, let $g^{\star}\in\arg\min_{g\in\mathcal{G}} \risk(g)$ and $\tilde g\in\arg\min_{g\in\mathcal{G}} \widetilde{\risk}(g;\hat\prior)$. Then
	\[
	\risk(\tilde g)-\risk(g^{\star})
	\;\le\; 2\,\sup_{g\in\mathcal{G}}\big|\widetilde{\risk}(g;\hat\prior)-\risk(g)\big|
	\;\le\; \frac{4B\,|\delta|}{\gamma^2}.
	\]
\end{theorem}
The full proof is given in Appendix C, see Theorem A.14.

\noindent Takeaways.
(i) Conditioning effect: the factor $\gamma^{-2}$ captures the ill-conditioning as $|\prior-\alpha|\!\to\!0$, explaining sensitivity to prior errors.  
(ii) Uniform control: for fixed $\alpha$, the bias scales linearly in $|\delta|$ and is uniformly bounded across $g$ for bounded $\phi$.

\medskip
\noindent Guidelines.
Design side: avoid regimes where $\alpha$ concentrates too closely around $\prior$ (e.g., mix tuples with distinct $(n,m)$ so that $\alpha$ varies).  
Estimation side: prefer prior estimators with small absolute error and, when possible, shrink $\hat\prior$ away from $\alpha$ if they fall on opposite sides.  
Mitigation: if $\phi$ is $L$-Lipschitz and $|g(\vx)|\le G$ a.s., replace $B$ by $LG$ in the above bounds.
\subsection{Misspecified Tuple Count (Random $m$)}\label{subsec:misspec-m}
The positive count $m$ may be noisy due to annotation errors or stochastic tuple construction; randomness in $m$ translates into randomness in $\alpha=m/n$, yielding a randomly conditioned inversion. We model the practitioner’s input as a random $\hat m\sim p(m')$ and study how the plug-in risk deviates from the truth.

\begin{theorem}[Risk bias under random count misspecification]\label{thm:count-misspec}
	Let $\vx\in\mathcal X$, $y\in\{+1,-1\}$ with prior $\prior\in(0,1)$, class-conditionals $\pp,\pn$, and unlabeled marginal $p_U=\prior\,\pp+(1-\prior)\,\pn$. For tuples of length $n$ with true count $m$ and $\alpha=m/n$,
	\[
	\pTtil \;=\; \alpha\,\pp + (1-\alpha)\,\pn, \qquad \text{assume }\prior\neq \alpha.
	\]
	Let $\ell(y,g(\vx))=\phi(y\,g(\vx))$ with $0\le\phi\le B$ and define $\risk(g)=\prior\,\E_{\pp}[\phi(g(\vx))]+(1-\prior)\,\E_{\pn}[\phi(-g(\vx))]$. Suppose, instead of $\alpha$, we use a random $\hat\alpha:=\hat m/n$ with $\hat m\sim p(m')$ and $\hat\alpha\neq \prior$ a.s. Given $(p_U,\pTtil)$, form
	\[
	\widehat p_{+}^{(\hat\alpha)}=\frac{(1-\hat\alpha)\,p_U-(1-\prior)\,\pTtil}{\prior-\hat\alpha},
	\qquad
	\widehat p_{-}^{(\hat\alpha)}=\frac{-\hat\alpha\,p_U+\prior\,\pTtil}{\prior-\hat\alpha},
	\]
	and the plug-in risk $\widetilde{\risk}(g;\hat\alpha)$ accordingly. Then, for every $g$ and every realization of $\hat\alpha$,
	\[
	\begin{split}
		\big|\widetilde{\risk}(g;\hat\alpha)-\,\risk(g)\big|
		\le
		\frac{2B\,\lvert\hat\alpha-\alpha\rvert}{\eta^{2}}, \\
		\eta:=\min_{\zeta\in[\min\{\alpha,\hat\alpha\},\,\max\{\alpha,\hat\alpha\}]}\!\lvert\prior-\zeta\rvert.
	\end{split}
	\]
	Taking expectation over $\hat\alpha$,
	\[
	\E_{\hat\alpha}\!\left[\big|\widetilde{\risk}(g;\hat\alpha)-\risk(g)\big|\right]
	\;\le\;
	\frac{2B}{\underline\eta^{\,2}}\;\E_{\hat\alpha}\!\left[\lvert\hat\alpha-\alpha\rvert\right],
	\underline\eta:=\inf_{\zeta\in\mathcal I}\!\lvert\prior-\zeta\rvert,
	\]
	where $\mathcal I$ contains (a.s.) the supports of $\alpha$ and $\hat\alpha$. Moreover, if $\bar \risk(g):=\E_{\hat\alpha}[\widetilde{\risk}(g;\hat\alpha)]$ and $\bar g\in\arg\min_{g\in\mathcal G}\bar \risk(g)$, then
	\[
	\risk(\bar g)-\risk(g^{\star})
	\;\le\; 2\sup_{g\in\mathcal G}\big|\bar \risk(g)-\risk(g)\big|
	\;\le\;\frac{4B}{\underline\eta^{\,2}}\;\E_{\hat\alpha}\!\left[\lvert\hat\alpha-\alpha\rvert\right].
	\]
\end{theorem}
The full proof is given in Appendix C, see Theorem A.15.

\begin{figure*}[t]
	\centering
	\scriptsize
	\includegraphics[scale=0.30]{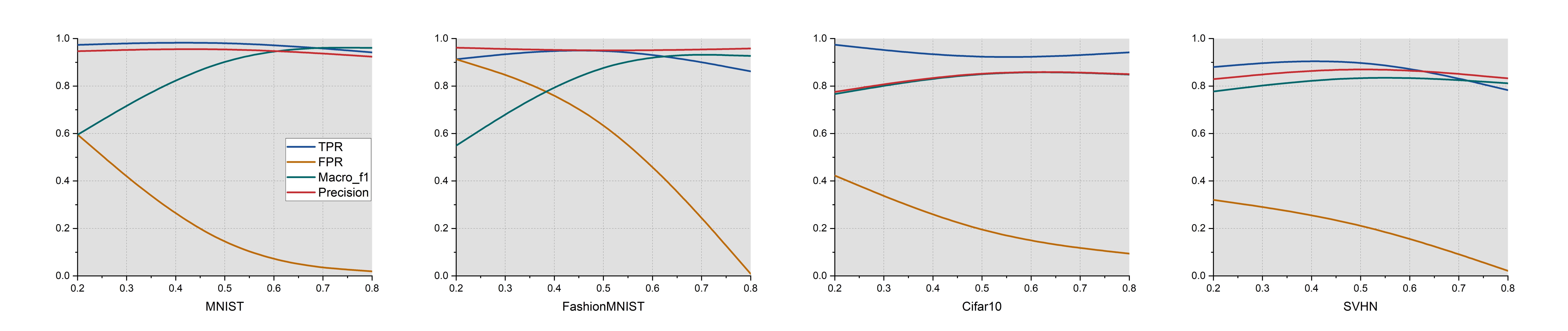}
	\caption{Tuples of length $n{=}3$ with $m{=}1$. Horizontal axis: class prior $\prior$; vertical axis: metric value. As $\prior$ increases, TPR, Macro-$F_{1}$, and Precision remain stable or improve across datasets, while FPR decreases markedly, indicating that prior shifts predominantly influence false positives.}
	\label{fig:measure}
\end{figure*}
\begin{figure*}[t]
	\centering
	\scriptsize
	\begin{tabular}{ccc}
		\includegraphics[width=3.6cm]{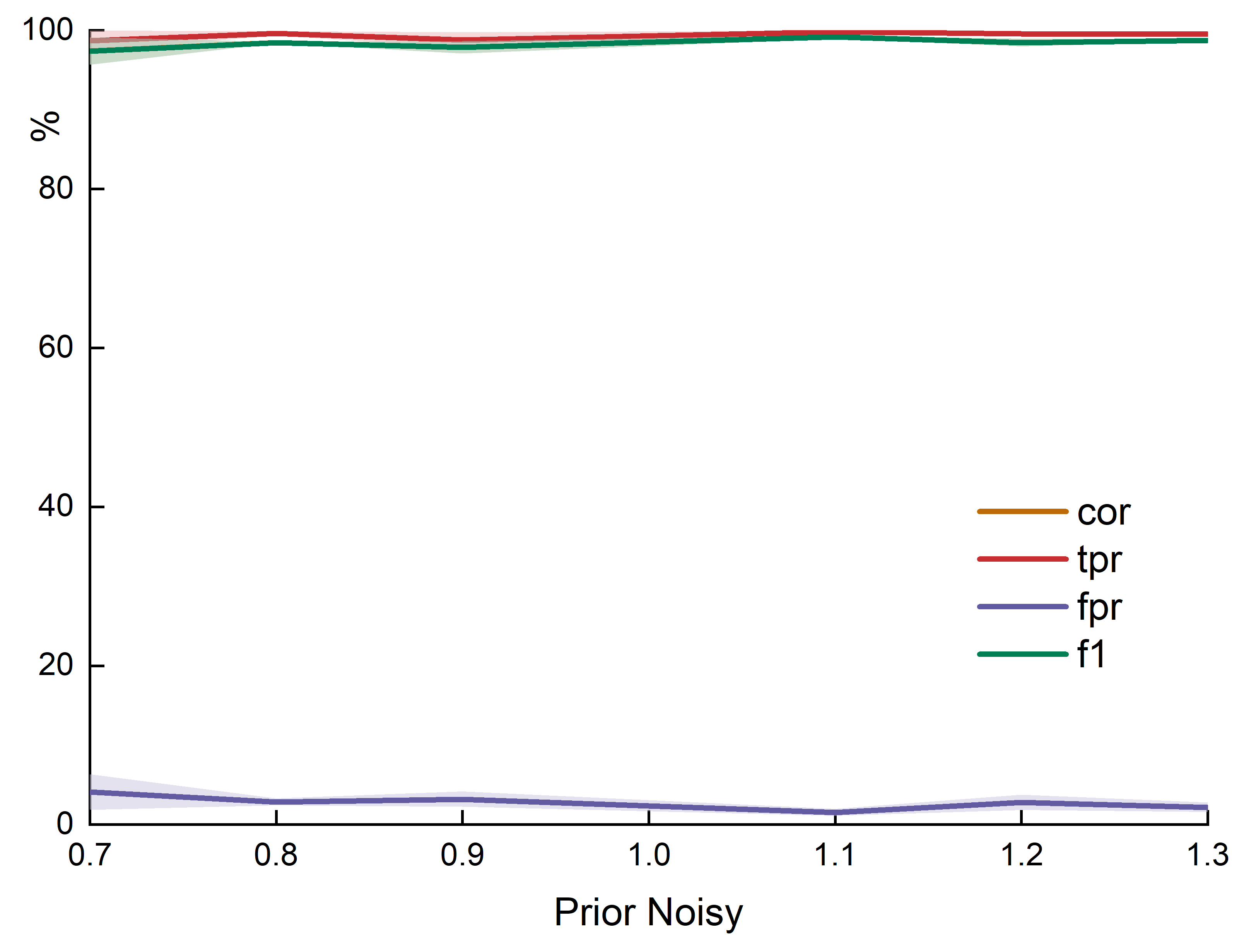} &
		\includegraphics[width=3.6cm]{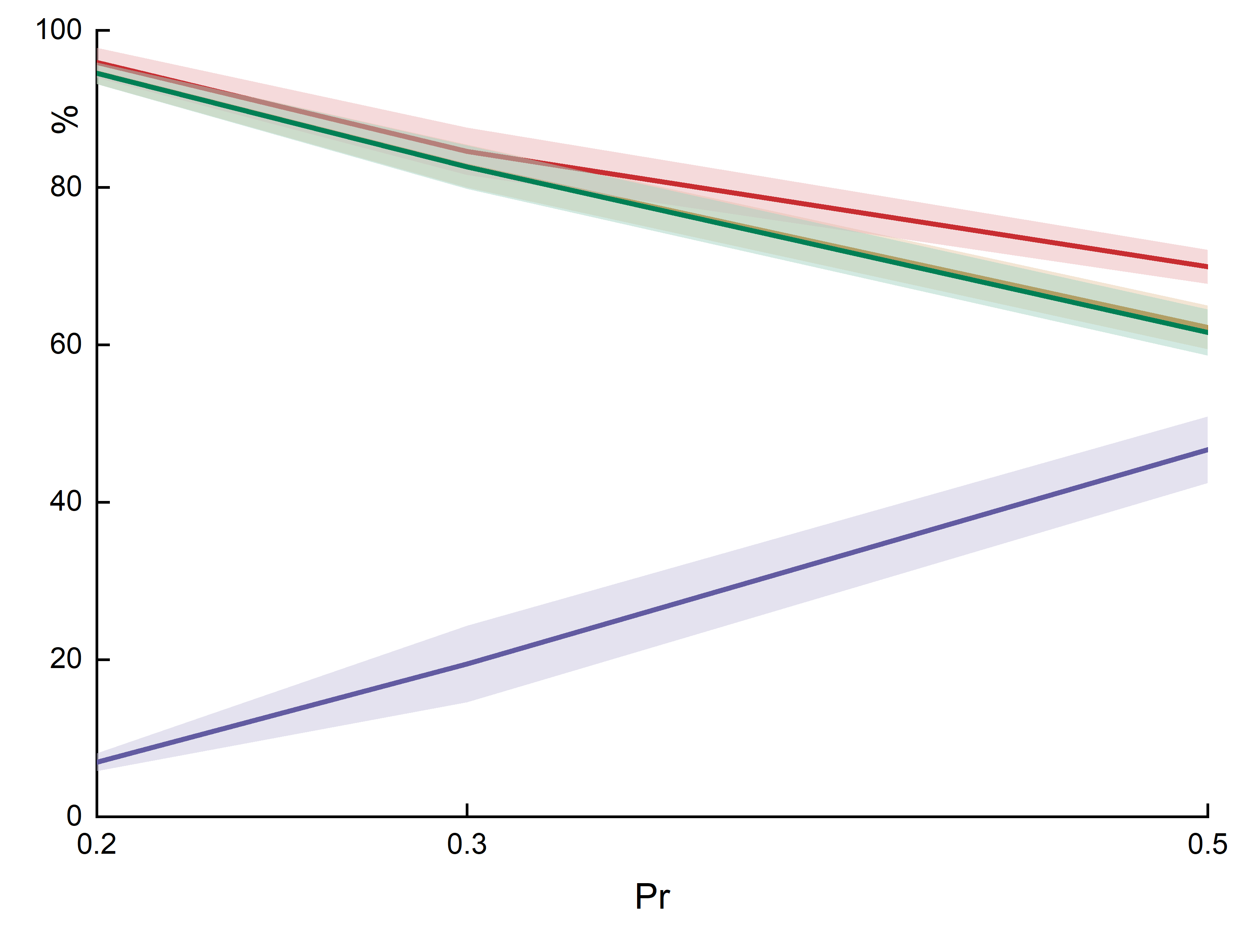} &
		\includegraphics[width=3.6cm]{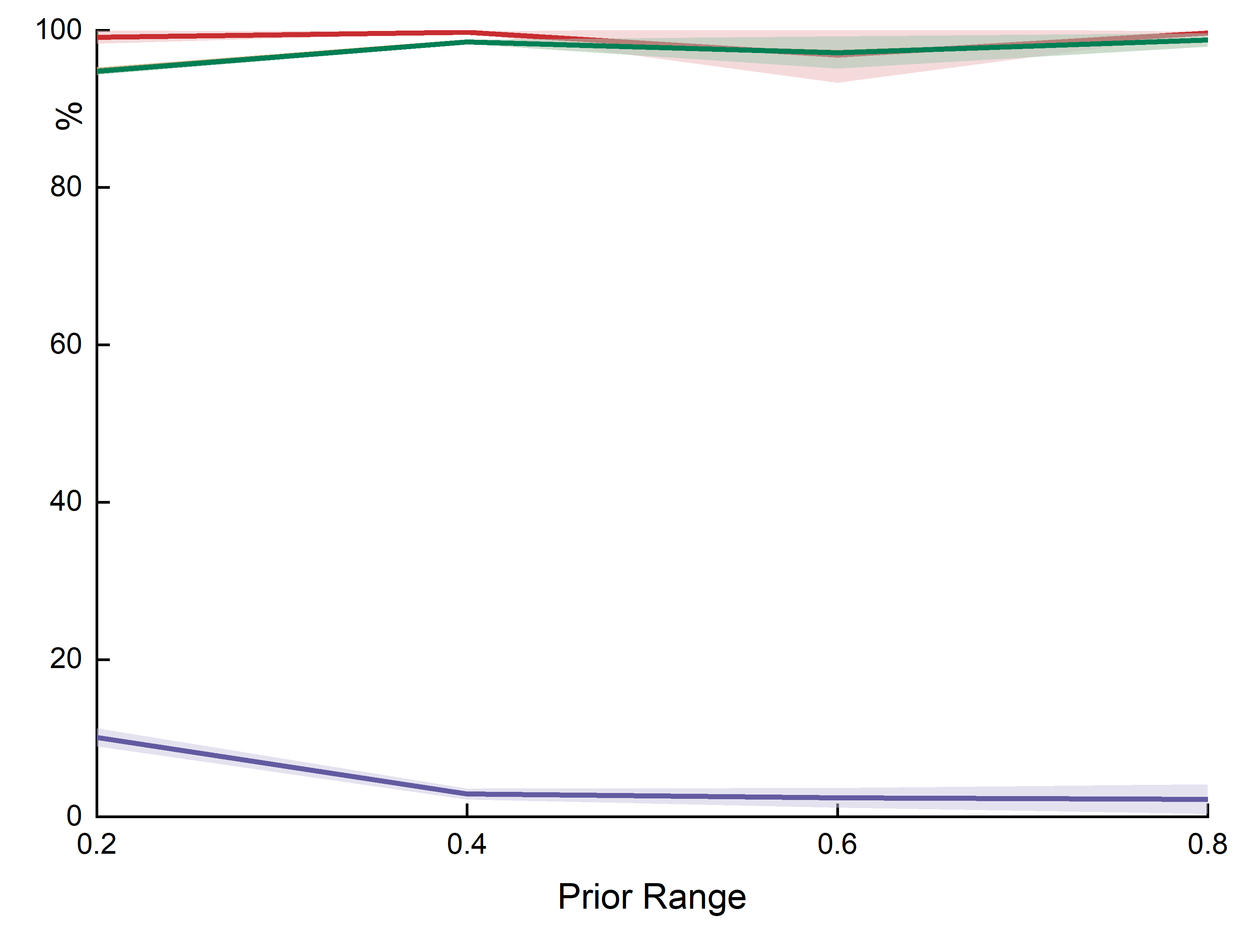} \\
		(a) \quad\;\; & (b) \quad\;\; & (c) \quad\;\;\\
	\end{tabular}
	\caption{Robustness \& conditioning on \textsc{MNIST}(\(S{=}3\) seeds; solid lines = mean; shaded = \(\pm\,1\) std).
		We report COR (accuracy), TPR, FPR, F1.
		(a) Prior sweep around the estimate: \(\dprior := \piTilde-\hat\pi \in [-0.30,0.30]\); COR/F1 change marginally and FPR stays near zero.
		\textbf{(b)} Tuple-count noise: with probability \(p\) each tuple is perturbed by \(m\mapsto m\pm1\); FPR rises roughly linearly while COR/TPR/F1 decrease monotonically.
		\textbf{(c)}  {Identifiability gap analysis:} sweep \(\Dalpha := |\pi-\alpha|\) by varying \(\pi\) over the shown range (fixing \(\alpha\)); metrics remain largely stable with small fluctuations and consistently low FPR, except in the immediate vicinity of \(\Dalpha\!\approx\!0\) (ill-conditioned boundary).}
	\label{fig:robust}
\end{figure*}
\begin{corollary}[Variance-form and high-probability variants]\label{cor:variance}
	Let $\sigma_\alpha^2:=\E\big[(\hat\alpha-\alpha)^2\big]$. Then $\E\lvert\hat\alpha-\alpha\rvert\le \sigma_\alpha$ and thus
	\[
	\sup_{g}\big|\bar \risk(g)-\risk(g)\big|
	\;\le\; \frac{2B}{\underline\eta^{\,2}}\;\sigma_\alpha,
	\qquad
	\risk(\bar g)-\risk(g^{\star})
	\;\le\; \frac{4B}{\underline\eta^{\,2}}\;\sigma_\alpha.
	\]
	If $\lvert\hat\alpha-\alpha\rvert\le \varepsilon$ a.s., then
	$\big|\widetilde{\risk}(g;\hat\alpha)-\risk(g)\big|\le \tfrac{2B\varepsilon}{\underline\eta^{\,2}}$ for all $g$.
	If $\hat\alpha-\alpha$ is sub-Gaussian with proxy variance $v^2$, then with probability at least $1-\delta$,
	\[
	\big|\widetilde{\risk}(g;\hat\alpha)-\risk(g)\big|
	\;\le\; \frac{2B}{\underline\eta^{\,2}}\; v\,\sqrt{2\log(1/\delta)}.
	\]
\end{corollary}

\noindent Remarks.
(i) The conditioning factor is $\lvert\prior-\hat\alpha\rvert^{-2}$, mirroring Theorem~\ref{thm:prior-misspec-bias}; the system is ill-conditioned as $\hat\alpha\!\to\!\prior$.  
(ii) Mixing tuples with diverse positive ratios enlarges $\underline\eta$, improving robustness.  
(iii) If $\phi$ is $L$-Lipschitz and $\lvert g(\vx)\rvert\le G$ a.s., replace $B$ by $LG$ throughout.
\begin{figure*}[t]
	\centering
	\scriptsize
	\begin{tabular}{cccc}
		\includegraphics[width=2.5cm]{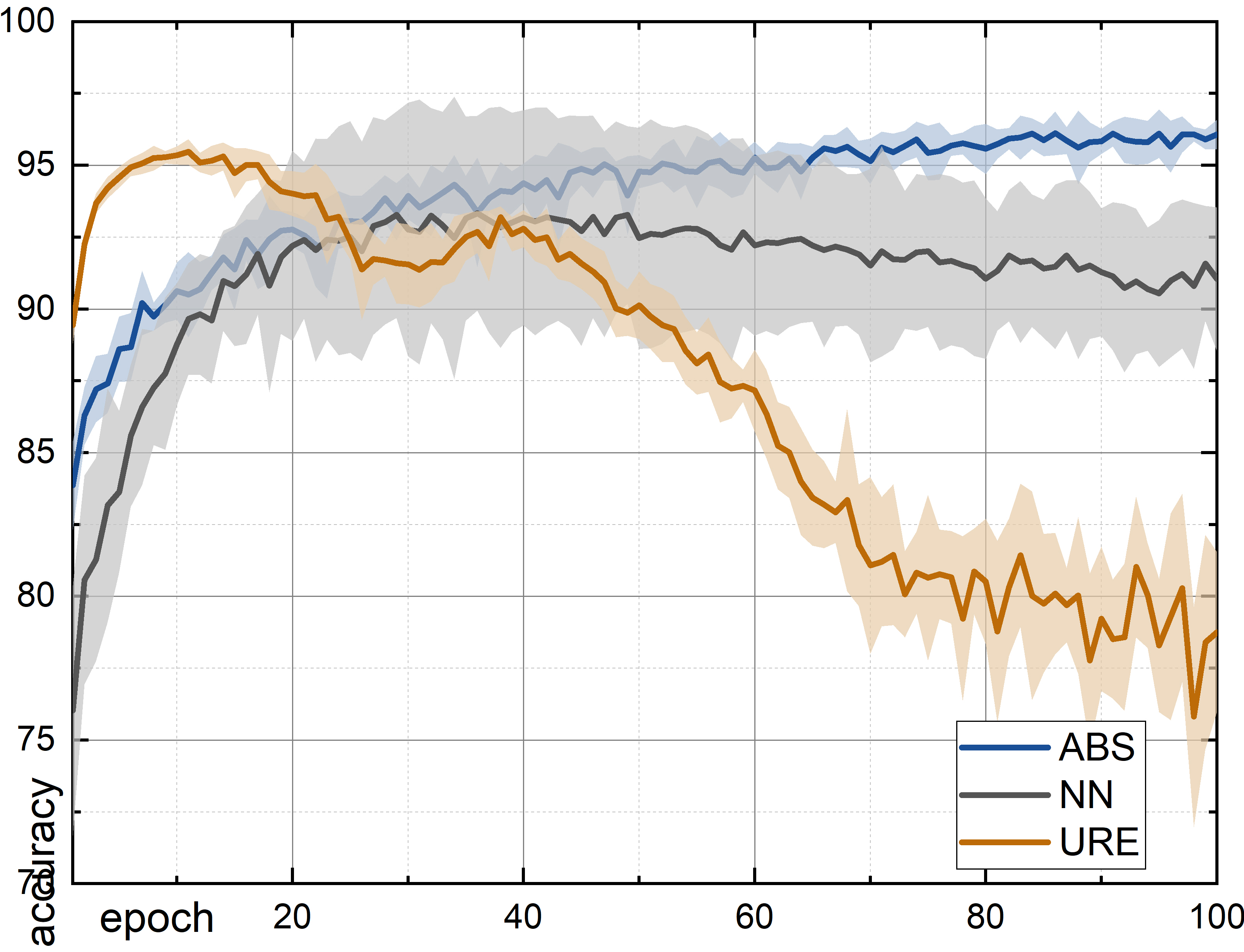} &
		\includegraphics[width=2.5cm]{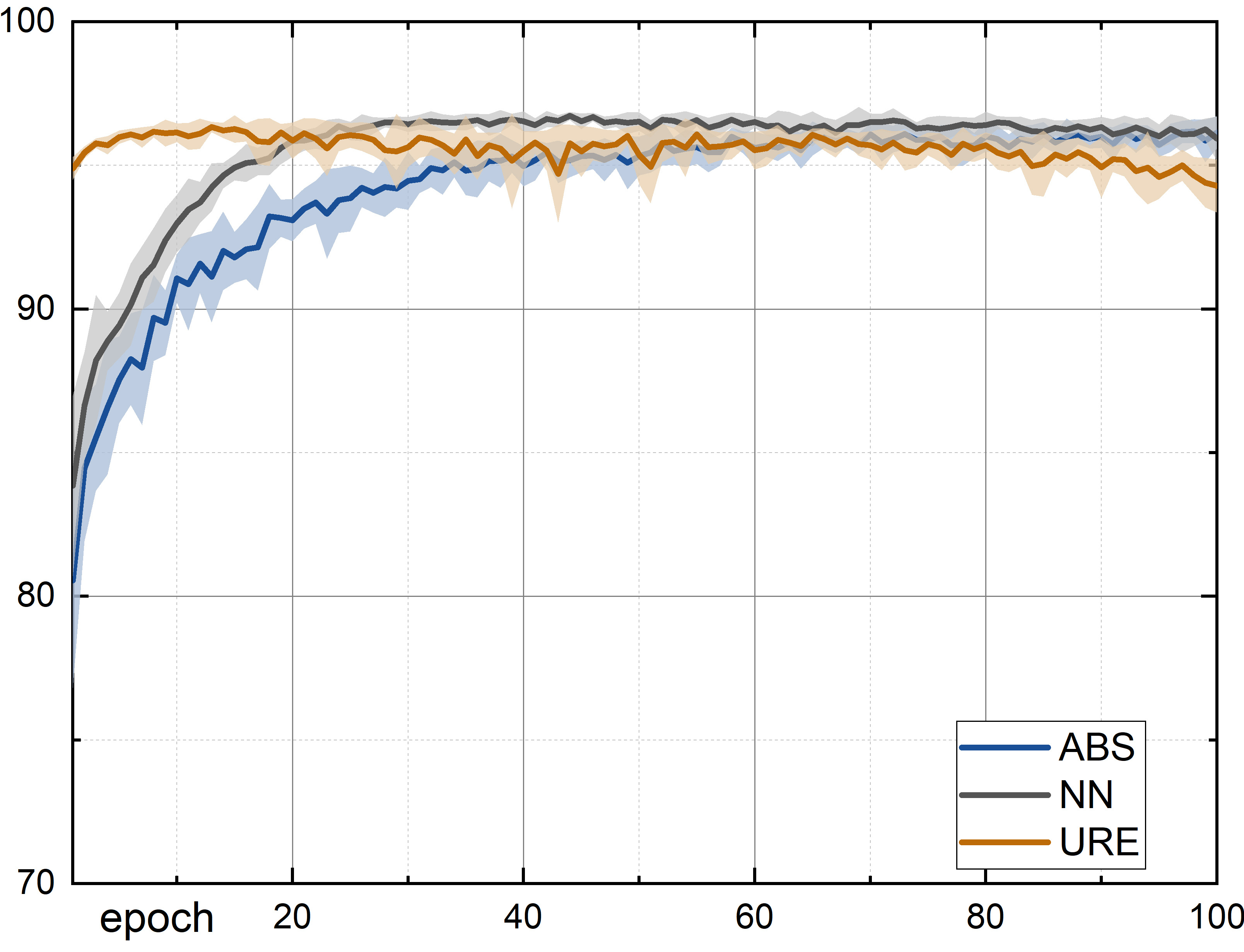} &
		\includegraphics[width=2.5cm]{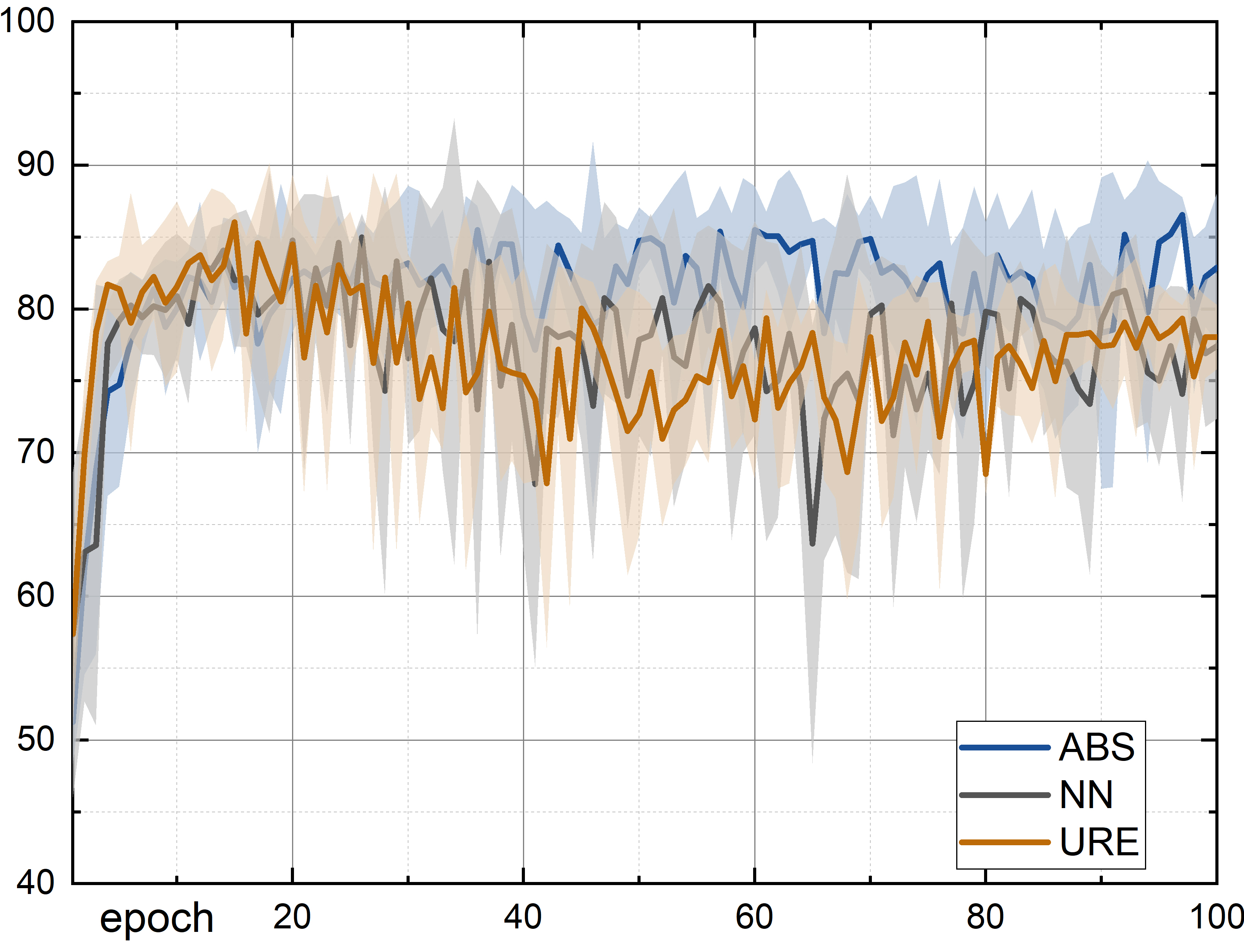} &
		\includegraphics[width=2.5cm]{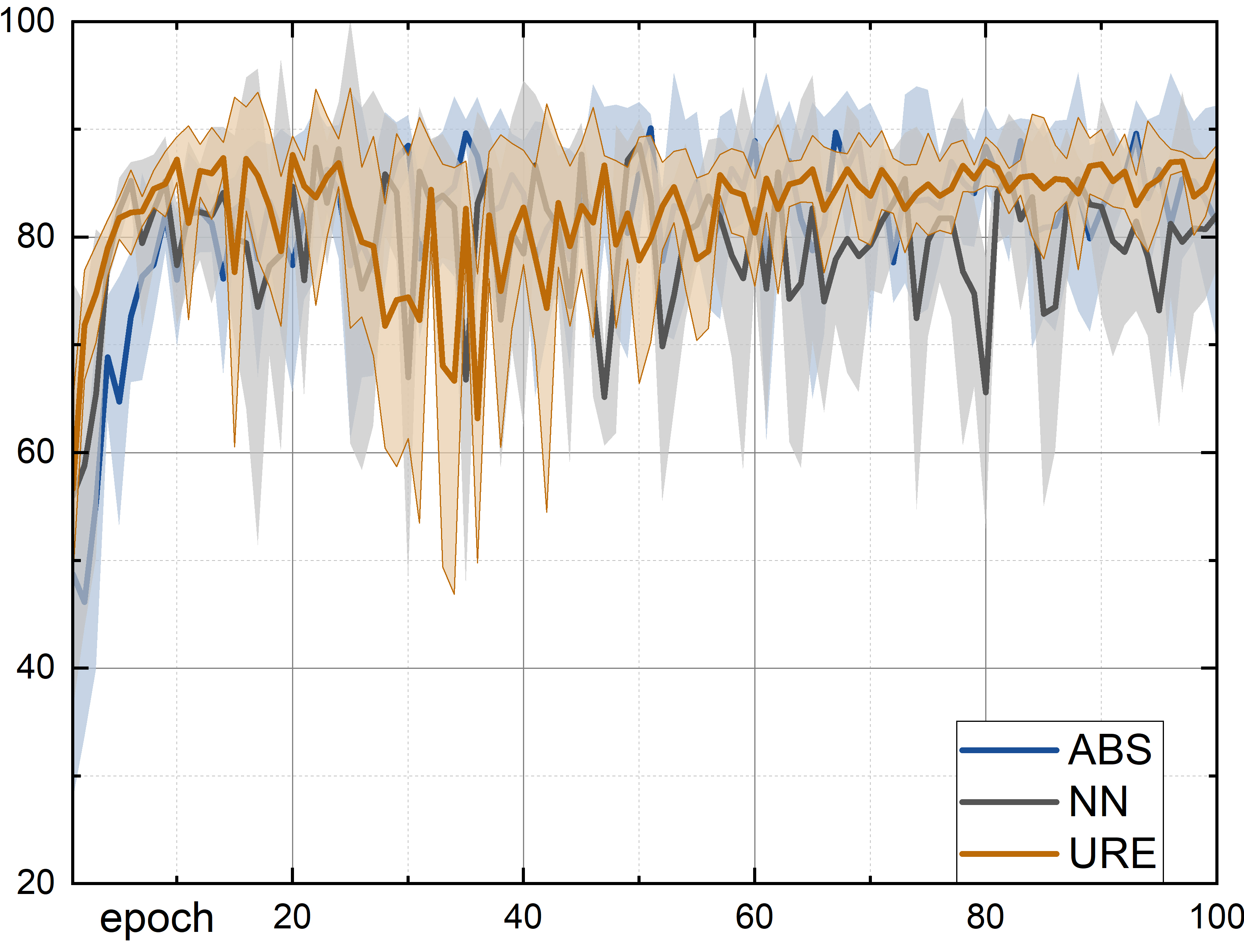} \\
		MNIST, MLP & \textsc{FashionMNIST}, MLP & CIFAR-10, ResNet \quad & SVHN, ResNet \\
	\end{tabular}
	\caption{Class prior $\pi\!=\!0.2$, tuples $(n,m)\!=\!(3,1)$. Horizontal axis: training iterations; vertical axis: accuracy. Mean $\pm$ std over five trials. The plain URE may overfit (accuracy spikes then decays), whereas the corrected variants stabilize training and improve the final accuracy (see text).}
	\vspace{-0.5em}
	\label{fig:compared}
\end{figure*}
\section{Experiments}\label{section5}
We empirically evaluate the proposed NTMP framework on four image benchmarks and compare it against relevant baselines. Our study addresses three questions:
(i) Does NTMP achieve higher accuracy than baselines under count-only supervision?
(ii) Do the stability corrections (ABS/ReLU clamps) mitigate overfitting?
(iii) How do the class prior $\pi$ and the tuple configuration $(n,m)$ affect COR/TPR/FPR/$F_{1}$ (and threshold–free metrics such as AP/AUROC)?
\subsection{Experimental Setup}
\paragraph*{Default prior regime and variants}
Unless otherwise stated, we adopt the known-by-construction prior regime:
the unlabeled pool $D_U$ is obtained by stripping labels from a disjoint training split,
so its class prior $\pi$ is known and plugged into the URE.
For robustness and the unknown-prior variant, we instead estimate a prior
$\hat{\pi}$ following Sec.~\ref{subsec:misspec-prior} (NP lower bound $\rightarrow$ MPE point estimate $\rightarrow$ $\Delta$-sweep),
and we report the operating gap $\Delta{=}\lvert \alpha{-}\hat{\pi}\rvert$ and the reproducible window $W(\Delta)$.

\paragraph*{Datasets and binary remappings}
Each dataset is converted to a binary task and then used to construct NTMP tuples. We report the specific $(n,m)$ (and the implied $\alpha{=}m/n$) in the captions of the corresponding tables/figures.
\textbf{MNIST}~\cite{726791} and \textsc{FashionMNIST}~\cite{xiao2017fashion} consist of $28{\times}28$ grayscale images
(60k/10k train/test), where for MNIST we treat even digits $\{0,2,4,6,8\}$ as positive and odd digits $\{1,3,5,7,9\}$ as negative, 
and for \textsc{FashionMNIST} we treat \{T-shirt/top, Pullover, Coat, Shirt, Bag\} as positive and 
\{Trouser, Dress, Sandal, Sneaker, Ankle boot\} as negative.
\textbf{SVHN}~\cite{2011Reading} contains $32{\times}32$ color digit images (73{,}257/26{,}032), with the same even-vs-odd
digit split as MNIST.
\textbf{CIFAR-10}~\cite{krizhevsky2009tinyimages} contains $32{\times}32$ color images (50k/10k), where we define the positive
class as \{airplane, bird, cat, dog, horse, truck\} and the negative class as \{automobile, deer, frog, ship\}.
Images are scaled to $[0,1]$. No augmentation is used unless specified.

\paragraph*{Tuple construction and unlabeled pool}\label{sec:tuple}
For each dataset, we construct $D_T$ by sampling $n_T$ tuples of fixed length $n\in\{3,5,7\}$ with an exact positive count $m\in\{1,2\}$ (so $\alpha=m/n$). Each tuple $T_i=(x_i^{(1)},\ldots,x_i^{(n)})$ is formed by drawing instances without replacement from the training split subject to the count constraint. We choose $n_T$ so that $n_T n$ matches the original training size (e.g., for MNIST, $n_T=60{,}000/n$). The auxiliary unlabeled set $D_U$ is created by stripping labels from a disjoint training pool; its class prior $\pi$ is therefore known by construction and is used in the URE terms (Sec.~\ref{section3}). For robustness analyses (prior–misspecification sweeps), we follow the protocol in Sec.~\ref{sec:prior-window} and report $S$-seed means with 95\% bootstrap CIs (captions specify $S$). Under the unknown-prior variant, we estimate $\hat{\pi}$ using the pipeline in Sec.~\ref{subsec:misspec-prior} (NP lower bound $\rightarrow$ MPE $\rightarrow$ $\Delta$-sweep) and then plug $\hat{\pi}$ into the same URE; unless otherwise stated, $\hat{\pi}$ is fixed per setup.

\paragraph*{Baselines and metrics}
We compare against representative weak-supervision baselines (UU and clustering-style surrogates), trained on the same backbone and data splits. Primary metrics are Macro-$F_{1}$, AP, and AUROC; we additionally report COR/TPR/FPR to diagnose error modes. All reported numbers are across $S$ seeds (default $S{=}5$) with 95\% bootstrap CIs (default $B{=}10{,}000$), matching the robustness protocol in Section~\ref{sec:prior-window}.

\paragraph*{Training and stability corrections}
Models are trained by minimizing the empirical URE in \eqref{eq:emp-risk} with mini-batches drawn from $D_T$ and $D_U$. As predicted by the theory (conditioning factor $\lvert\pi{-}\alpha\rvert^{-1}$ in \eqref{eq:gen-bound}), variance inflates near $\alpha{\approx}\pi$. We therefore report results both with and without the nonnegative clamp (ABS/ReLU) described in Section~\ref{section3-b}; the clamp stabilizes finite-sample behavior without altering the population optimum of $\risk(g)$.

\paragraph*{Reporting of $(n,m)$, prior sweep, and gaps}
To tie experiments to the misspecification/conditioning theory (Sec.~\ref{subsec:misspec-prior}–\ref{subsec:misspec-m}), each main table/figure reports:
(i) the operating gap $\Dop:=|\alpha-\hat\pi|$ (aggregated across $S$ seeds);
(ii) when a prior sweep is performed, curves versus the  {signed offset} $\dprior=\piTilde-\hat\pi$, with the robustness window $W_{\dprior}(\varepsilon,w^\star)$ shaded and the critical point(s) $\delta_{\text{crit}}^{\text{CI}}$ (and $\delta_{\text{crit}}^{\text{perf}}$ when used) marked;
(iii) the exact tuple configuration(s) $(n,m)$ used (for heterogeneous tuples we also report the effective rate $\alphabar$ and stratum sizes).
When conditioning is analyzed, we additionally display the  {identifiability gap} $\Dalpha:=|\pi-\alpha|$ explored in that ablation.
\paragraph*{Metrics and statistical tests}
We report average precision (AP), area under the ROC curve (AUROC), and $F_1$-score as main performance metrics. For calibration, we use the bin-based expected calibration error (ECE) and the Brier score (binary and multiclass); unless otherwise stated, we adopt $M{=}15$ equal-width bins and standard definitions (full formulas are given in Appendix~B).

We additionally report temperature-scaled variants $\ECE^{\mathrm{TS}}$ and $\Brier^{\mathrm{TS}}$, where a scalar temperature $T{>}0$ is fitted on a held-out validation set by minimizing the negative log-likelihood and then applied to the test set.

For significance testing, we use a paired Wilcoxon signed-rank test comparing each method to \textsc{Ours}, and adjust $p$-values using Holm's step-down procedure; we denote the adjusted values by $\pHolm(\AP)$, $\pHolm(\AUROC)$, etc. Effect sizes are summarized by Cliff's delta on the paired per-seed differences (with the usual negligible/small/medium/large interpretation; see Appendix~B). Unless noted, all curves and tables show mean $\pm$ 95\% bootstrap confidence intervals over seeds.

\begin{table*}[t]
	\centering\small
	\caption{Main results on \textsc{FashionMNIST} ($S{=}5$ seeds). Mean$\pm$std; 95\% CI in brackets.
		$p$-values are Holm-adjusted paired Wilcoxon vs.~\textsc{URE}; $\Cliff$ is Cliff's $\delta$.}
	\label{tab:fmnist_main}
	\begin{tabular}{r l c c c c c c}
		\hline
		$\Delta$ & Method & AP$\uparrow$ & AUROC$\uparrow$ & ECE\textsubscript{TS}$\downarrow$ & Brier\textsubscript{TS}$\downarrow$ & $p_{\text{Holm}}$(AP) & Cliff's $\delta$(AP) \\
		\hline
		0.00 & ABS & 0.99$\pm$0.008[0.002]
		
		& 0.99$\pm$0.008[0.002] & 0.11[0.005]
		& 0.06[0.012] & \textbf{0.000} & \textbf{0.815} \\
		0.00 & URE  & 0.77$\pm$0.279[0.086]
		
		& 0.69$\pm$0.382[0.117] & 0.29[0.090]
		& 0.27[0.090] & -- & -- \\
		0.25 & ABS & 0.99$\pm$0.005[0.002]
		
		& 0.99$\pm$0.003 [0.001] & 0.09[0.007]
		& 0.05[0.008] & \textbf{0.000} & \textbf{0.331} \\
		0.25 & URE  & 0.36$\pm$0.033[0.019]
		& 0.14$\pm$0.142[0.073] & 0.59[0.120]
		& 0.52[0.127] & -- & -- \\
		\hline
	\end{tabular}
	\label{tab:perf-calib-sig}
\end{table*}
\subsection{Models and Training Protocol}\label{subsec:models}
\textsc{MNIST/FashionMNIST.} A 4-layer MLP (300 units per layer, ReLU, BatchNorm \cite{ioffe2015batch}) producing a scalar score $g(\vx)$. Optimizer: Adam, learning rate $10^{-4}$, $100$ epochs.
\textsc{SVHN/CIFAR-10.} A lightweight ResNet-20 (9 residual blocks), Adam with learning rate $5\!\times\!10^{-4}$, $100$ epochs; standard per-channel normalization. No heavy augmentation is used to isolate supervision effects.

\textbf{Optimization of URE.} We minimize the empirical unbiased risk \eqref{eq:emp-risk} over mixed mini-batches (tuples from $D_T$ and samples from $D_U$). Unless otherwise noted, we use uniform in-tuple averaging (Lemma~\ref{lem:variance}) and optionally apply a nonnegative clamp (ABS/ReLU) inside the tuple/unlabeled terms (Sec.~\ref{section3-b}) to stabilize training. As predicted by the conditioning factor $\lvert\pi-\alpha\rvert^{-1}$, we additionally report results both with and without clamping when $\alpha\!\approx\!\pi$.
\subsection{Baselines}\label{subsec:baselines}
We compare NTMP against methods that operate under weak/aggregate supervision.

\paragraph{K-Means (KM)}
Unsupervised clustering with $k{=}2$ on raw (or penultimate) features.
Clusters are mapped to labels by matching cluster proportions to the known $\pi$ via a two-way assignment (equivalent to majority mapping when $\pi{=}0.5$).
This provides a purely unsupervised reference.

\paragraph{UU learning}
Risk reconstruction from two unlabeled sets with distinct priors \cite{lu2018minimal,lu2020mitigating}.
We instantiate one set as $D_U$ (prior $\pi$) and the other as the flattened tuple pool $\tilde D_T$ (effective prior $\alpha$), yielding a direct UU objective on the same backbone. UUcor denotes UU unbiased risk with absolute corrected function.

\paragraph{Clustering+Classifier}
Cluster first, then train a linear classifier on cluster pseudo-labels (self-training style).
This probes whether count signals in NTMP offer gains beyond structure-only assignments.

\paragraph{Attention-MIL (non-LLM)}
A representative MIL baseline with attention pooling \cite{ilse2018attention}.
Each bag (tuple) $\mathcal{B}=\{x_i\}_{i=1}^{L}$ is encoded by a frozen image encoder into $z_i\in\R^{d}$; an attention head produces
$a_i=\mathrm{softmax}_i(w^\top\tanh(W z_i))$, and the bag embedding
$h=\sum_i a_i\,\tanh(W z_i)$ feeds a linear classifier
$s_{\text{bag}}=v^\top h$.
We train only the attention and linear heads with a bag-level binary loss, using identical backbones, budgets, and seeds as NTMP.
For instance-level analysis, we also report per-instance margins $s_i=v^\top \tanh(W z_i)$ without instance labels.

\paragraph{CLIP-MIL (LLM-assisted)}
A lightweight LLM-assisted MIL variant using a frozen CLIP encoder.
Let $f(\cdot)$ and $g(\cdot)$ be CLIP image/text encoders, with normalized embeddings $\hat f(x)$ and $\hat g(t)$.
The instance margin is the cosine score $s_i=\langle \hat f(x_i), \hat g(t)\rangle$ for a class prompt; the bag score uses a permutation-invariant aggregator (mean pooling):
$s_{\text{bag}}=\frac{1}{L}\sum_i s_i$.
No additional head is fitted; supervision remains strictly at the bag level.

\paragraph{Bag-level Cross-Entropy Proportion Matching (LLP)}\label{bagece}
For each bag, average instance softmax outputs to obtain $\bar{\boldsymbol p}$ and match to the target proportions $\boldsymbol q$ via $H(\boldsymbol q,\bar{\boldsymbol p})$; a light instance-entropy penalty stabilizes training.
This is a standard, strong LLP baseline.

\paragraph{Distribution Matching with Jensen--Shannon Divergence (LLP)}\label{dmjs}
Same setup as above but replace the bag-level cross-entropy with the symmetric, bounded $\mathrm{JS}(\bar{\boldsymbol p}\,\|\,\boldsymbol q)$, which typically improves robustness under noisy or slightly misspecified bag proportions while keeping supervision at the bag level.

\begin{table}[!t]
	\centering
	\caption{Small-scale comparison on MNIST tuples under count-only supervision.
		We compare attention-based MIL (no LLM), CLIP-MIL (LLM prompts), and two LLP baselines—
		Bag-level Cross-Entropy Proportion Matching (BagCE) and Distribution Matching with Jensen--Shannon divergence (DM-JS).
		All encoders are frozen for fairness and only bag-level objectives are optimized.
		Numbers are mean$\pm$std over 3 runs; $\uparrow$ means higher is better.
		NTMP (URE) attains the strongest AP/Best-F1/AUC, and the simple ABS correction further stabilizes and improves performance.}
	\label{tab:llm_mil_small}
	\begin{tabular}{lccc}
		\hline
		Method & AP $\uparrow$ & Best-F1 $\uparrow$ & AUC $\uparrow$  \\
		\hline
		Attention-MIL (no LLM) & 0.85$\pm$0.01 & 0.80$\pm$0.00 & 0.91$\pm$0.00  \\
		CLIP-MIL (LLM prompts) & 0.25$\pm$0.00 & 0.57$\pm$0.00 & 0.13$\pm$0.00 \\
		LLP–BagCE & 0.37$\pm$0.00 & 0.67$\pm$0.00 & 0.26$\pm$0.01 \\
		LLP–JS & 0.38$\pm$0.00 & 0.67$\pm$0.00 & 0.27$\pm$0.01 \\
		\hline
		NTMP (URE)       & 0.92$\pm$0.01 & 0.84$\pm$0.02 & 0.86$\pm$0.02   \\
		\quad + ABS correction  & \textbf{0.99$\pm$0.00} & \textbf{0.98$\pm$0.00} & \textbf{0.99$\pm$0.00}  \\
		\hline
	\end{tabular}
\end{table}
\begin{table*}[hpbt]
	\begin{center}
		\caption{Under the same experimental settings, the results presents the performance of different methods across four datasets under the same experimental setup, where the length of each tuple is set to 5 and each tuple contains 2 positive instances.}
		\label{table:flfp2}
		\setlength{\tabcolsep}{1pt}
		\begin{tabular}{ccccccccc}
			\hline
			\multirow{2}{*}{prior} & 
			\multirow{2}{*}{Datasets} & 
			\multicolumn{3}{l}{$P_{NTMP}U$} & 
			\multicolumn{4}{l}{\multirow{1}{*}{$Baseline$}} \\
			
			& & ABS & NN & URE & UUcor & \multicolumn{1}{c}{UU} & KM & KM++\\
			\hline
			\multirow{4}{*}{0.2} & MNIST & \textbf{96.15$\pm$0.17} & 95.93$\pm$0.23 & 94.28$\pm$0.28 & 94.85$\pm$0.27 & 78.27$\pm$1.95 & 61.09$\pm$0.03 & 61.09$\pm$0.01 \\
			& \textsc{FashionMNIST} & \textbf{94.68$\pm$0.24} & 91.97$\pm$0.24 & 78.67$\pm$0.86 & 93.14$\pm$0.10 & 94.46$\pm$0.47 & 67.78$\pm$0.01  & 67.89$\pm$0.03 \\
			& CIFAR10 & \textbf{82.89$\pm$1.47} & 77.43$\pm$1.43 & 78.06$\pm$0.64& 76.53$\pm$0.63 & 71.41$\pm$1.38 & 51.59$\pm$0.00 & 51.39$\pm$0.01 \\
			& SVHN & \textbf{84.35$\pm$1.35} & 65.07$\pm$0.76 & 83.59$\pm$1.68 & 85.25$\pm$1.13 & 78.84$\pm$1.20 & 55.87$\pm$0.00  & 55.80$\pm$0.00 \\
			\hline
			\multirow{4}{*}{0.5} & MNIST & 97.76$\pm$0.41 & \textbf{98.27$\pm$0.28} & 81.13$\pm$1.11 & 61.44$\pm$0.60 & 61.52$\pm$1.10 & 59.21$\pm$0.03  & 59.21$\pm$0.01   \\
			& \textsc{FashionMNIST} & 94.08$\pm$0.65 & \textbf{96.82$\pm$0.14} & 65.39$\pm$0.50 & 66.53$\pm$0.92 & 61.67$\pm$0.53 & 68.52$\pm$0.00  & 68.54$\pm$0.00  \\
			& CIFAR10 & \textbf{82.87$\pm$1.06} & 68.16$\pm$3.96 & 77.83$\pm$0.87& 56.72$\pm$1.34 & 56.09$\pm$0.74 & 51.21$\pm$0.00    & 51.11$\pm$0.00  \\
			& SVHN & \textbf{89.12$\pm$1.04} & 82.41$\pm$2.05 & 81.87$\pm$1.35 & 53.49$\pm$1.65 & 51.02$\pm$0.00 & 55.42$\pm$0.00   & 55.45$\pm$0.00   \\
			\hline
			\multirow{4}{*}{0.8} & MNIST & 95.80$\pm$0.17 & \textbf{96.21$\pm$0.17} & 92.89$\pm$0.31 & 92.29$\pm$0.35 & 88.28$\pm$0.50 & 67.40$\pm$0.03  & 67.40$\pm$0.00  \\
			& \textsc{FashionMNIST} & \textbf{92.40$\pm$0.26} & 91.31$\pm$0.34 & 86.75$\pm$0.31 & 91.78$\pm$0.21 & 83.49$\pm$0.17 & 64.57$\pm$0.00    &  64.69$\pm$0.00 \\
			& CIFAR10 & \textbf{81.68$\pm$2.08} & 74.73$\pm$1.61 & 79.54$\pm$0.59 & 77.08$\pm$0.45 & 68.32$\pm$0.74  & 52.66$\pm$0.00  & 52.71$\pm$0.02 \\
			& SVHN & 80.83$\pm$1.58 & \textbf{83.66$\pm$0.67} & 78.71$\pm$1.39 & 81.90$\pm$0.43 & 75.32$\pm$6.95 & 57.47$\pm$0.01   & 57.68$\pm$0.00  \\
			\hline
		\end{tabular}
	\end{center}
\end{table*}	
\begin{table*}[hpbt]
	\begin{center}
		\caption{Under the same experimental settings, we set the length of each tuple to 7, with each tuple containing exactly two positive instance.}
		\label{table:flfp3}
		\setlength{\tabcolsep}{2pt}
		\begin{tabular}{ccccccccc}
			\hline
			\multirow{2}{*}{prior} & 
			\multirow{2}{*}{Datasets} & 
			\multicolumn{3}{l}{$P_{NTMP}U$} & 
			\multicolumn{4}{l}{\multirow{1}{*}{$Baseline$}} \\
			
			& & ABS & NN & URE & UUcor & \multicolumn{1}{c}{UU} & KM & KM++\\
			\hline		
			\multirow{4}{*}{0.2} & MNIST & 95.55$\pm$0.12 & \textbf{96.23$\pm$0.15} & 95.75$\pm$0.15 & 92.92$\pm$0.59 & 87.86$\pm$0.40 & 63.99$\pm$0.00 & 65.44$\pm$0.01 \\
			& \textsc{FashionMNIST} & 93.62$\pm$0.21 & \textbf{94.32$\pm$0.16} & 87.31$\pm$0.41 & 92.34$\pm$0.30 & 92.69$\pm$0.13 & 68.73$\pm$0.00 & 68.91$\pm$0.00 \\
			& CIFAR10 & 74.82$\pm$3.82 & \textbf{77.79$\pm$1.94} & 71.79$\pm$2.73 & 76.34$\pm$1.09 & 76.38$\pm$0.70 & 51.12$\pm$0.00 & 52.64$\pm$0.00 \\
			& SVHN & 72.67$\pm$2.18 & \textbf{92.78$\pm$0.70} & 90.15$\pm$1.31 & 85.20$\pm$0.95 & 86.10$\pm$2.16 & 51.33$\pm$0.00 & 51.29$\pm$0.01 \\
			\hline 
			\multirow{4}{*}{0.5} & MNIST & 95.87$\pm$0.40 & \textbf{96.96$\pm$0.32} & 84.57$\pm$1.55 & 52.88$\pm$0.46 & 59.20$\pm$0.60 & 64.14$\pm$ 0.01   & 65.68$\pm$0.01   \\
			& \textsc{FashionMNIST} & 87.96$\pm$0.76 & \textbf{92.76$\pm$0.41} & 60.10$\pm$0.74 & 51.62$\pm$0.45 & 49.66$\pm$1.26  & 68.36$\pm$0.00    &  68.15$\pm$0.01  \\
			& CIFAR10 & \textbf{77.90$\pm$2.38} & 71.92$\pm$3.29 & 76.47$\pm$2.18 & 58.42$\pm$2.69 & 57.80$\pm$0.94 & 52.44$\pm$0.00   & 52.62$\pm$0.01   \\
			& SVHN & 81.99$\pm$1.15 & 80.07$\pm$2.74 & \textbf{86.96$\pm$2.07} & 53.73$\pm$1.29 & 49.73$\pm$0.94 & 51.59$\pm$0.04   & 51.72$\pm$0.02  \\
			\hline
			\multirow{4}{*}{0.8} & MNIST & \textbf{96.01$\pm$0.10} & 96.23$\pm$0.12 & 93.03$\pm$0.37 & 93.66$\pm$0.50 & 86.72$\pm$0.77 & 63.99$\pm$0.00  & 63.52$\pm$0.01   \\
			& \textsc{FashionMNIST} & \textbf{93.71$\pm$0.16} & 92.95$\pm$0.31 & 87.20$\pm$0.23 & 91.26$\pm$0.07 & 82.84$\pm$0.08 & 70.88$\pm$0.03   & 71.17$\pm$0.01   \\
			& CIFAR10 & 75.70$\pm$4.02 & 77.29$\pm$0.77 & \textbf{80.58$\pm$0.53} & 74.80$\pm$1.20 & 69.85$\pm$0.65 & 51.59$\pm$0.00    & 51.78$\pm$0.01   \\
			& SVHN & 80.57$\pm$0.91 & \textbf{84.91$\pm$1.84} & 75.27$\pm$3.35 & 83.59$\pm$1.56 & 75.78$\pm$6.83 & 50.46$\pm$0.01  & 50.32$\pm$0.04   \\	
			\hline
		\end{tabular}
	\end{center}
\end{table*}
\begin{figure*}[t]
	\centering
	\scriptsize
	\begin{tabular}{cccc}
		\includegraphics[width=2.8cm]{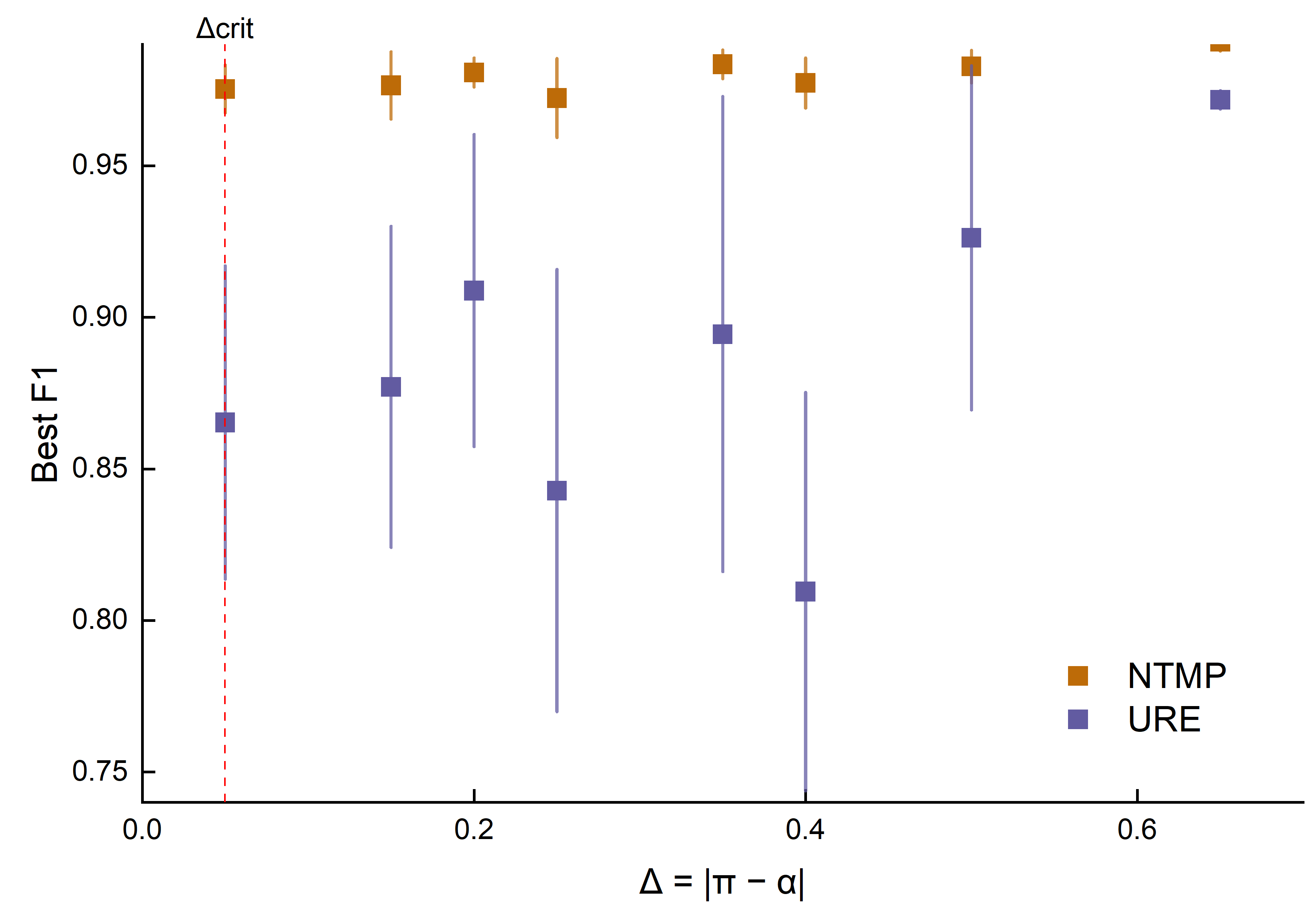} &
		\includegraphics[width=2.8cm]{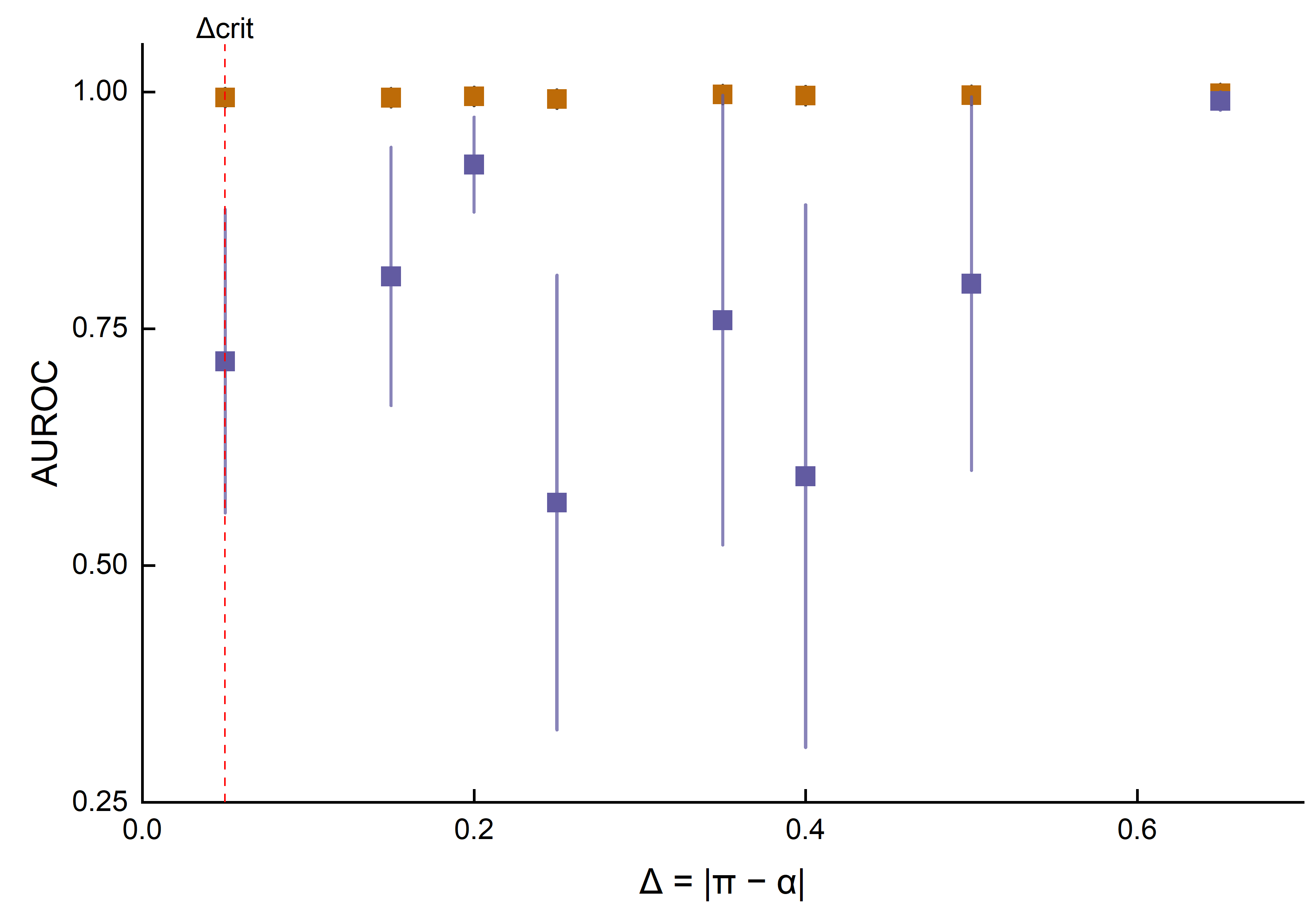} &
		\includegraphics[width=2.8cm]{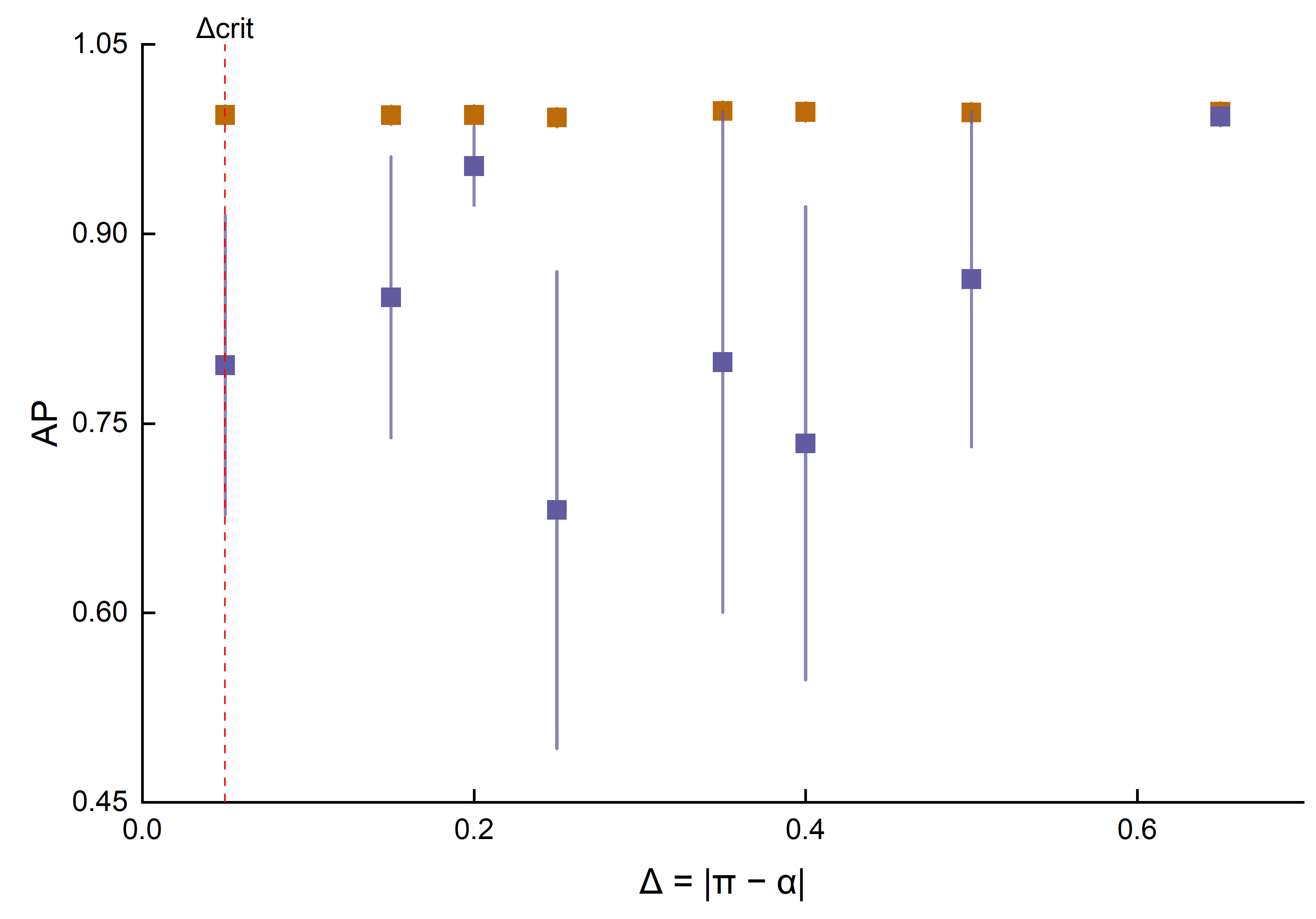} &
		\includegraphics[width=2.8cm]{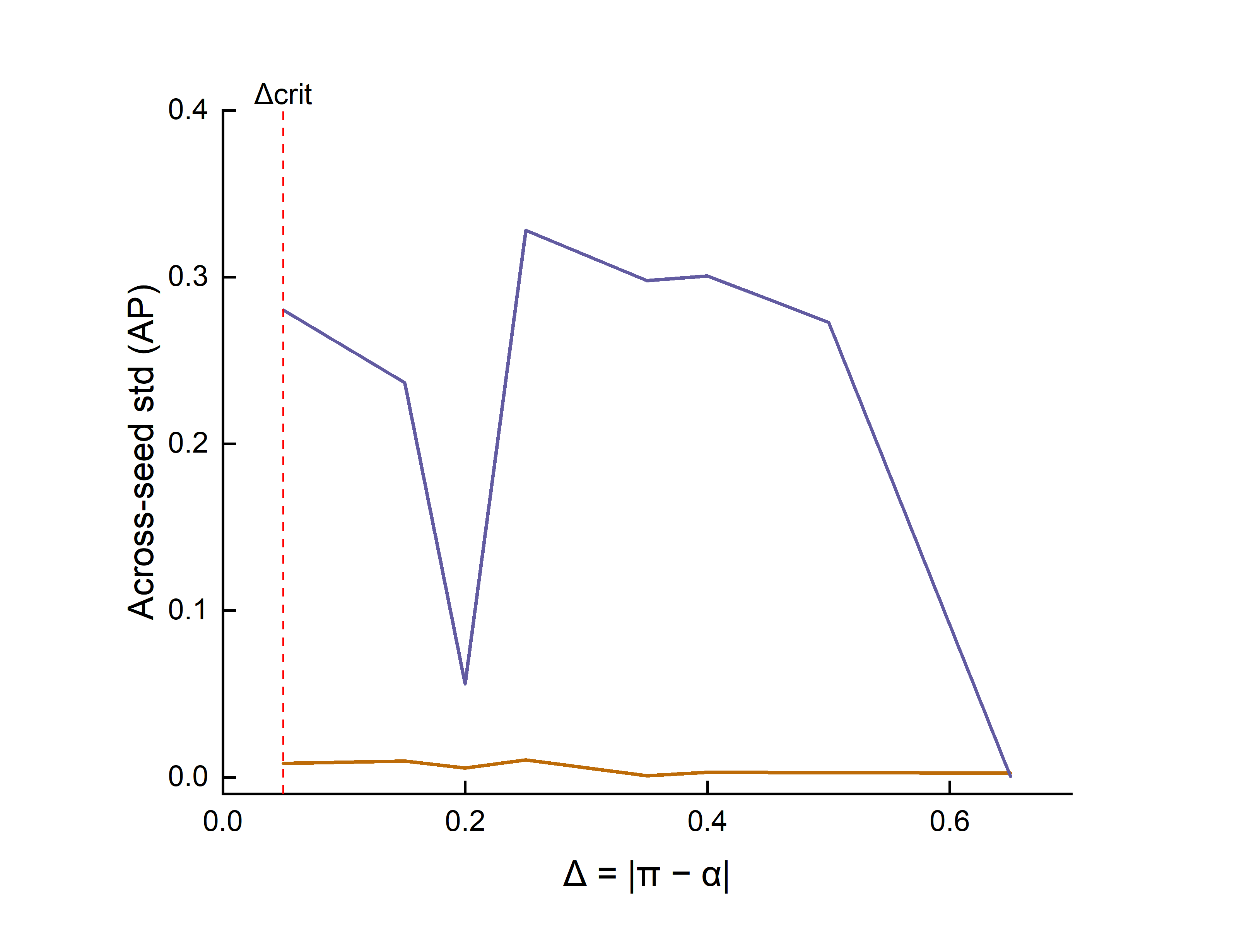} \\
		(a) & (b) & (c) & (d) \\
	\end{tabular}
		\caption{Metrics and stability under identifiability gap (\(\Dalpha\)) on MNIST.
		Horizontal axis: \(\Dalpha:=|\pi-\alpha|\), where \(\pi\) is the prior used in training for that run and \(\alpha=m/n\) is the tuple positive rate of the \((n,m)\) configuration.
		We compare URE+ABS against URE.
		Markers show mean over \(S{=}3\) seeds; vertical bars denote 95\% bootstrap CIs.
		The red dashed line indicates the CI–based critical gap
		\({\Dalpha}_{\text{crit}}^{\text{CI}}\), defined as the smallest \(\Dalpha\) at which the 95\% CI width of AP first exceeds the preset threshold \(w^\star{=}0.05\).
		(a) Best-\(F_1\) vs.\ \(\Dalpha\).
		(b) AUROC vs.\ \(\Dalpha\).
		(c) AP vs.\ \(\Dalpha\).
		(d) Stability vs.\ \(\Dalpha\), measured by across-seed std of AP (larger is less stable).
		Across the \(\Dalpha\)-sweep, URE+ABS maintains high accuracy with near-zero dispersion, whereas plain URE degrades and becomes increasingly variable as the prior–tuple mismatch grows.}
	\label{fig:delta}
\end{figure*}

\subsection{Results}\label{subsec:results}

We first compare test accuracy across methods. Table~\ref{table:flfp1} summarizes results for $(n,m)=(3,1)$ under three unlabeled-set priors $\pi\in\{0.2,0.5,0.8\}$, with each entry reported as mean$\pm$std over five independent tuple constructions. Taken together with Table~\ref{tab:perf-calib-sig}, NTMP(+ABS) not only improves AP/AUROC but also reduces ECE\textsubscript{TS}/Brier\textsubscript{TS}, with small Holm–adjusted $p$-values and positive Cliff’s $\delta$ (small–to–medium), indicating statistically significant and practically meaningful gains.

Both corrected variants (ABS/ReLU) consistently lead, often by large margins. For example, on \textsc{CIFAR-10} with $\pi{=}0.5$, NTMP(ABS) attains $\approx 80.0\%$ vs.\ $67.8\%$ for UUcor and $69.2\%$ for plain URE. On \textsc{MNIST} and \textsc{FashionMNIST}, NTMP frequently exceeds $95\%$, whereas UU degrades notably when $\pi\neq\alpha$, highlighting the benefit of exploiting exact tuple counts rather than treating $\tilde D_T$ as a generic unlabeled set.

In a few cases (e.g., \textsc{SVHN} at $\pi{=}0.2$), URE matches or slightly exceeds corrected NTMP, confirming that the unbiased objective carries signal. However, URE shows higher variance and overfitting, while ABS/ReLU provide steadier gains across datasets and priors—consistent with our variance and conditioning analysis.

Adding a clamp improves UU over its uncorrected form, yet the gap to NTMP remains substantial—especially around $\pi{=}0.5$—suggesting that count-aware risk reconstruction outperforms mixing two unlabeled bags that ignore tuple constraints.

KM/KM++ hover around $50$--$60\%$ in balanced regimes and rarely exceed trivial baselines when imbalanced, confirming that count-only supervision provides signal that unsupervised clustering fails to extract.

We repeat the comparison for $n\in\{5,7\}$ (with appropriate $m$); see Tables~\ref{table:flfp2}--\ref{table:flfp3}. Trends persist: NTMP leads consistently, and the advantage grows with larger $n$ (weaker per-instance signal), where exact counts become more valuable. We observe mild, dataset-dependent differences between ABS and ReLU: for $(n,m)=(7,2)$ ABS sometimes edges out ReLU, whereas for $(3,1)$ ReLU may slightly win; both effectively reduce overfitting.

Table~\ref{tab:llm_mil_small} compares count-only supervision on MNIST tuples with frozen encoders. NTMP (URE) surpasses the MIL and LLP baselines (AP $0.92\!\pm\!0.01$, Best-$F_{1}$ $0.84\!\pm\!0.02$, AUROC $0.86\!\pm\!0.02$). Adding ABS further lifts performance to near ceiling (AP $0.99\!\pm\!0.00$, Best-$F_{1}$ $0.98\!\pm\!0.00$, AUROC $0.99\!\pm\!0.00$). Attention–MIL is competitive (AP $0.85\!\pm\!0.01$, Best-$F_{1}$ $0.80\!\pm\!0.00$, AUROC $0.91\!\pm\!0.00$), whereas the LLM-assisted CLIP–MIL variant lags under these settings. LLP baselines (BagCE/DM-JS) provide weaker references. Variances are uniformly small across three runs, indicating a stable ranking under a fixed compute/data protocol.

Figure~\ref{fig:delta} summarizes the identifiability-gap sweep
$(\Dalpha:=|\pi-\alpha|)$. Across the range, URE+ABS attains
near-saturated Best-$F_{1}$, AUROC, and AP with almost zero across-seed variance.
The red dashed line marks the CI-based critical gap ${\Dalpha}_{\text{crit}}^{\text{CI}}$.
In contrast, plain URE degrades as $\Dalpha$ increases and exhibits much
larger dispersion (AP across-seed std typically $0.2$–$0.3$ in the mid-range).
These patterns are consistent over the $(n,m)$ settings used in this figure.
Paired Wilcoxon tests on AP favor URE+ABS at most $\Dalpha$ values
(significance markers $*, **, ***$).

Figure~\ref{fig:compared} (\(\pi{=}0.2\), $(n,m){=}(3,1)$) shows that URE achieves high training accuracy quickly and then fluctuates—typical overfitting—whereas clamped objectives stabilize training and improve final test accuracy.
Figure~\ref{fig:measure} reports TPR/FPR/Precision/Macro-$F_{1}$ as $\pi$ varies at fixed $(3,1)$: TPR and Macro-$F_{1}$ are stable or improve with larger $\pi$, and FPR decreases markedly, indicating robustness to prior shifts.
Figure~\ref{fig:robust} evaluates three perturbations on \textsc{MNIST}:
(a) prior misspecification up to $\pm30\%$ leaves COR/$F_{1}$ essentially flat and keeps FPR near zero;
(b) under synthetic count noise (flip $m$ to $m\!\pm\!1$ with probability $\Pr$), FPR increases approximately linearly in $\Pr$, while COR/TPR/$F_{1}$ decrease monotonically;
(c) sweeping $\pi$ across the near-degenerate regime ($\alpha\!\approx\!\pi$) yields small fluctuations except in a narrow band around the boundary.
All observations match the theoretical predictions of Sec.~\ref{subsec:misspec-prior}--\ref{subsec:misspec-m}. Across the three stressors (prior misspecification, count flips, and \(\alpha\!\approx\!\pi\)), COR/TPR/\(F_{1}\) remain stable outside a narrow ill-conditioned band, and FPR grows smoothly under count noise—consistent with the \(\lvert\pi-\alpha\rvert^{-1}\) sensitivity predicted by our theory (Fig.~\ref{fig:robust}).

NTMP is preferable when exact counts \((n,m)\) are available and tuples can be treated as orderless mixtures; the observable pair (tuple aggregate, unlabeled pool) closes a \(2{\times}2\) system identifiable whenever the effective rate \(\bar{\alpha}\neq\pi\).
LLP requires sufficient cross-bag proportion variation—near-identical proportions render the problem ill-posed and high-variance.
MIL is appropriate when intra-bag dependencies are discriminative (stable spatial/temporal structure, sparse yet clustered positives, and weak per-instance evidence requiring neighborhood/positional context); it tends to overfit when instances are i.i.d.\ within a bag, whereas NTMP may underperform if critical relational cues are lost under mixture flattening.
\begin{table}[hpbt]
	\begin{center}
		\caption{Accuracy on four datasets with $(n,m)=(3,1)$ and class prior $\pi\in\{0.2,0.5,0.8\}$.
			Values are mean$\pm$std over five runs. ABS/ReLU denote nonnegative clamps applied inside
			the tuple and unlabeled terms of the URE.}
		\label{table:flfp1}
		\setlength{\tabcolsep}{2pt}
		\begin{tabular}{ccccccccc} 
			\hline
			\multirow{2}{*}{prior} & 
			\multirow{2}{*}{Datasets} & 
			\multicolumn{3}{l}{$P_{NTMP}U$} & 
			\multicolumn{4}{l}{\multirow{1}{*}{$Baseline$}} \\
			
			& & ABS & NN & URE & UUcor & \multicolumn{1}{c}{UU} & KM & KM++\\
			\hline
			\multirow{4}{*}{0.2} & MNIST & 95.50$\pm$0.09 & \textbf{96.44$\pm$0.05} & 94.90$\pm$0.25 & 92.94$\pm$0.32 & 85.92$\pm$0.81 & 58.44$\pm$0.03  & 58.52$\pm$0.01 \\
			& \textsc{FashionMNIST} & 93.08$\pm$0.33 & \textbf{93.57$\pm$0.20} & 84.51$\pm$0.54 & 92.60$\pm$0.12 & 91.30$\pm$0.44 & 73.11$\pm$0.01 & 73.15$\pm$0.01 \\
			& CIFAR10 & \textbf{80.94$\pm$1.49} & 72.20$\pm$3.07 & 73.29$\pm$1.56 & 76.08$\pm$0.45 & 69.37$\pm$1.60 & 50.80$\pm$0.00  & 50.63$\pm$0.00 \\
			& SVHN & 79.77$\pm$1.80 & 87.15$\pm$1.45 & \textbf{87.57$\pm$0.84} & 71.92$\pm$2.37 & 82.82$\pm$0.46& 52.34$\pm$0.00  & 52.14$\pm$0.00\\
			\hline
			\multirow{4}{*}{0.5} & MNIST & 97.76$\pm$0.44 & \textbf{98.87$\pm$0.18} & 87.53$\pm$0.46 & 68.84$\pm$0.61 & 61.88$\pm$0.33 & 60.73$\pm$0.01   & 59.96$\pm$0.01   \\
			& \textsc{FashionMNIST} & 93.45$\pm$0.67 & \textbf{96.39$\pm$0.55} & 78.50$\pm$0.73 & 73.83$\pm$1.03 & 73.01$\pm$0.00 & 73.48$\pm$0.00  & 73.42$\pm$0.00   \\
			& CIFAR10 & \textbf{80.01$\pm$2.11} & 73.99$\pm$2.61 & 69.17$\pm$1.50 & 67.80$\pm$1.39 & 62.24$\pm$0.77 & 51.12$\pm$0.01  &  51.04$\pm$0.00  \\
			& SVHN & 78.11$\pm$1.83 & \textbf{86.12$\pm$1.91} & 83.66$\pm$0.72 & 59.90$\pm$2.29 & 50.22$\pm$5.85& 51.83$\pm$0.00  & 51.84$\pm$0.00   \\
			\hline
			\multirow{4}{*}{0.8} & MNIST & 98.98$\pm$0.17 & \textbf{99.07$\pm$0.20} & 94.88$\pm$0.25 & 87.25$\pm$0.51 & 73.90$\pm$0.14 & 51.33$\pm$0.01  & 51.41$\pm$0.01   \\
			& \textsc{FashionMNIST} & 93.47$\pm$0.74 & \textbf{96.21$\pm$0.33} & 82.20$\pm$0.65 & 89.77$\pm$1.28 & 73.27$\pm$0.10 & 69.72$\pm$0.04   & 70.84$\pm$0.03   \\
			& CIFAR10 & \textbf{79.43$\pm$1.27} & 73.90$\pm$1.91 & 66.91$\pm$2.22 & 74.21$\pm$0.93 & 69.03$\pm$0.56 & 50.29$\pm$0.01   & 50.34$\pm$0.03   \\
			& SVHN & \textbf{84.25$\pm$2.52} & 80.65$\pm$2.04 & 65.61$\pm$3.65 & 77.98$\pm$3.50 & 62.56$\pm$8.96 & 53.75$\pm$0.00   & 54.07$\pm$0.00   \\
			\hline
		\end{tabular}
	\end{center}
\end{table}
\section{Conclusion}\label{section6}
We studied weakly supervised learning from NTMP and showed how to train a binary classifier without instance labels. Our key contribution is a closed-form URE that couples the flattened tuple mixture $\pTtil$ (with weight $\alpha{=}m/n$) and an auxiliary unlabeled pool $p_U$ (with prior $\pi$), eliminating the unknown class-conditionals via a $2{\times}2$ linear system. On top of this objective, we established generalization guarantees with optimal $N^{-1/2}$ rates (up to logs), characterized identifiability and conditioning through the factor $\lvert \pi-\alpha\rvert^{-1}$, and derived robustness results for prior misspecification and random count noise (uniform bias bounds and excess-risk control). We also introduced simple nonnegative stability corrections (ABS/ReLU) that reduce variance while preserving asymptotic correctness.

Empirically, NTMP consistently matches or outperforms representative weak-supervision alternatives (e.g., UU variants and clustering) and maintains favorable COR/TPR/FPR/$F_{1}$ trade-offs. The observed trends align with theory: performance remains stable under class-prior shifts and tuple reconfigurations, with degradation confined to the narrow ill-conditioned regime $\alpha\!\approx\!\pi$ predicted by the conditioning analysis.

Limitations and future work.
(i) Flattening tuples for risk estimation may forgo informative intra-tuple structure; extending NTMP to tuple-aware architectures under count supervision is a natural direction.
(ii) Multi-class and hierarchical-count extensions would broaden applicability.
(iii) Jointly learning prior and count calibrations (e.g., EM or auxiliary calibration losses) could further improve robustness when $\pi$ and $m$ are uncertain.
(iv) Beyond simple clamps, adaptive clipping or control variates may tighten finite-sample variance for URE.

Overall, NTMP shows that coarse count supervision, combined with an unlabeled reference and principled risk reconstruction, is a scalable, theoretically grounded, and practically stable alternative where instance-level annotation is costly or infeasible.
\bibliographystyle{elsarticle-num}
\bibliography{ref}

\end{document}